\documentclass{article}

\usepackage[nonatbib, final]{neurips_2022}

\usepackage[utf8]{inputenc} %
\usepackage[T1]{fontenc}    %
\usepackage{url}            %
\usepackage{booktabs}       %
\usepackage{amsfonts}       %
\usepackage{nicefrac}       %
\usepackage{microtype}      %

\usepackage{color,xcolor}
\usepackage{epsfig}
\usepackage{graphicx}
\usepackage{subfiles}

\usepackage{adjustbox}
\usepackage{array}
\usepackage{booktabs}
\usepackage{colortbl}
\usepackage{float,wrapfig}
\usepackage{hhline}
\usepackage{multirow}
\usepackage{subcaption} %
\usepackage[labelfont={bf},labelsep={period},font={small}]{caption}

\usepackage{tabu}

\let\llncssubparagraph\subparagraph
\let\subparagraph\paragraph
\usepackage[compact]{titlesec}
\let\subparagraph\llncssubparagraph

\usepackage{amsmath,amsfonts,amssymb}
\usepackage{bm}
\usepackage{nicefrac}
\usepackage{microtype}
\usepackage{mathtools}

\usepackage{changepage}
\usepackage{extramarks}
\usepackage{fancyhdr}
\usepackage{lastpage}
\usepackage{setspace}
\usepackage{soul}
\usepackage{xspace}

\usepackage{url}

\usepackage{algorithm}
\usepackage{algpseudocode}
\usepackage{enumerate}

\usepackage{times}
 
\usepackage{pbox}
\usepackage{footnote}
\usepackage{tablefootnote}

\usepackage{paralist}

\usepackage{enumitem}

\definecolor{mydarkgreen}{rgb}{0.02,0.6,0.02}
\usepackage[colorlinks, citecolor=mydarkgreen]{hyperref}       %
\usepackage{pifont}

\usepackage[normalem]{ulem}

\usepackage{ dsfont }

\usepackage{pifont}
\usepackage{nccmath}

\usepackage{paralist}
\usepackage{listings}
\usepackage{xcolor}

\definecolor{codegreen}{rgb}{0,0.6,0}
\definecolor{codegray}{rgb}{0.5,0.5,0.5}
\definecolor{codepurple}{rgb}{0.58,0,0.82}
\definecolor{backcolour}{rgb}{0.95,0.95,0.92}

\definecolor{mydarkblue}{rgb}{0,0.08,1}
\definecolor{mydarkred}{rgb}{0.8,0.02,0.02}
\definecolor{mydarkorange}{rgb}{0.40,0.2,0.02}
\definecolor{mypurple}{RGB}{111,0,255}
\definecolor{myred}{rgb}{1.0,0.0,0.0}
\definecolor{mygold}{rgb}{0.75,0.6,0.12}
\definecolor{mydarkgray}{rgb}{0.66, 0.66, 0.66}
\definecolor{mygray}{gray}{0.9}

\lstdefinestyle{mystyle}{
    backgroundcolor=\color{backcolour},   
    commentstyle=\color{codegreen},
    keywordstyle=\color{magenta},
    numberstyle=\tiny\color{codegray},
    stringstyle=\color{codepurple},
    basicstyle=\ttfamily\footnotesize,
    breakatwhitespace=false,         
    breaklines=true,                 
    captionpos=b,                    
    keepspaces=true,                 
    numbers=left,                    
    numbersep=5pt,                  
    showspaces=false,                
    showstringspaces=false,
    showtabs=false,                  
    tabsize=2
}
\usepackage{hyperref}

\hypersetup{%
    colorlinks=true,
    urlcolor=blue,
}

\definecolor{mydarkblue}{rgb}{0,0.08,1}

\newcommand{\myparagraph}[1]{\vspace{0pt}\paragraph{#1}}

\newcommand{\ignorethis}[1]{}

\makeatletter
\DeclareRobustCommand\onedot{\futurelet\@let@token\@onedot}
\def\@onedot{\ifx\@let@token.\else.\null\fi\xspace}

\def\eg{\emph{e.g}\onedot} 
\def\ie{\emph{i.e}\onedot} 
 
\def\etc{\emph{etc}\onedot} \def\vs{\emph{vs}\onedot}

\makeatother

\makeatletter

\newcommand\footnoteref[1]{\protected@xdef\@thefnmark{\ref{#1}}\@footnotemark}
\makeatother

\definecolor{fig5gray}{rgb}{0.80, 0.85, 0.89}
\definecolor{fig5red}{rgb}{1.00, 0.59, 0.55}
\definecolor{fig5yellow}{rgb}{1.00, 0.68, 0.00}

\def\engine{Tiny Training Engine\xspace}
\def\engineshort{TTE\xspace}

\title{
On-Device Training Under 256KB Memory}

\author{
Ji Lin$^{1*}$
\space%
Ligeng Zhu$^{1*}$
\space
Wei-Ming Chen$^1$
\space
Wei-Chen Wang$^1$
\space
Chuang Gan$^2$
\space
Song Han$^1$ \\
$^1$MIT \quad $^2$MIT-IBM Watson AI Lab  \\
\url{https://tinyml.mit.edu}
}

\begin{document}

\maketitle

\footnotetext{$*$ indicates equal contributions.}

\begin{abstract}

On-device training enables the model to adapt to new data collected from the sensors by fine-tuning a pre-trained model. Users can benefit from customized AI models without having to transfer the data to the cloud, protecting the privacy. However, the training memory consumption is prohibitive for IoT devices that have tiny memory resources.
We propose an algorithm-system co-design framework
to make on-device training possible with only \emph{256KB} of memory. On-device training faces two unique challenges: (1) the quantized graphs of neural networks are hard to optimize due to low bit-precision and the lack of normalization; (2) the limited hardware resource does not allow full back-propagation.
To cope with the optimization difficulty,
we propose \textit{\textbf{Quantization-Aware Scaling}}  to calibrate the gradient scales and stabilize 8-bit quantized training.
To reduce the memory footprint, we propose \textit{\textbf{Sparse Update}} to skip the gradient computation of less important layers and sub-tensors.
The algorithm innovation is implemented by a lightweight training system, \textit{\textbf{\engine}}, which prunes the backward computation graph to support sparse updates and offload the runtime auto-differentiation to compile time.
Our framework is the \emph{first} solution to enable tiny on-device training of convolutional neural networks under 256KB SRAM and 1MB Flash without auxiliary memory, using less than 1/1000 of the memory of PyTorch and TensorFlow while matching the accuracy on tinyML application VWW~\cite{chowdhery2019visual}. Our study enables IoT devices not only to perform inference but also to continuously adapt to new data for on-device lifelong learning. A video demo can be found \href{https://youtu.be/0pUFZYdoMY8}{here}.

\end{abstract}

\vspace{-5pt}
\section{Introduction}

On-device training allows us to \emph{adapt} the pre-trained model to newly collected sensory data \emph{after} deployment.  By training and adapting \emph{locally} on the edge, the model can learn to improve its predictions and perform lifelong learning and user customization. For example, fine-tuning a language model enables continual learning from users' typing and writing; adapting a vision model enables recognizing new objects from a mobile camera. By bringing training closer to the sensors, it also helps to protect user privacy when handling sensitive data (\eg, healthcare). 

\begin{figure}
    \centering
    \includegraphics[width=\textwidth]{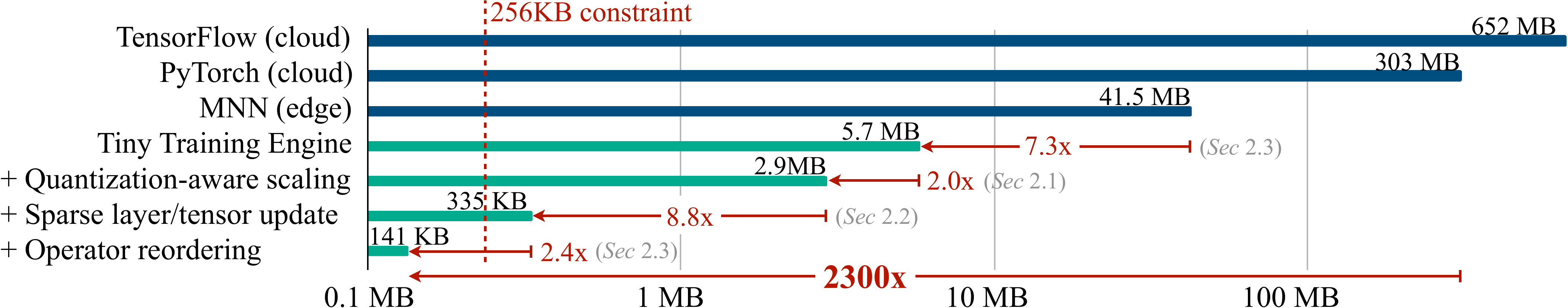}
    \caption{Algorithm and system co-design reduces the training memory from 303MB (PyTorch) to 141KB with the same transfer learning accuracy, leading to 2300$\times$ reduction. The numbers are measured with MobilenetV2-w0.35~\cite{sandler2018mobilenetv2}, batch size 1 and resolution 128$\times$128. It can be deployed to a microcontroller with 256KB SRAM.
    }
    \label{fig:teaser}
\end{figure}

However, on-device training on tiny edge devices is extremely challenging and fundamentally different from cloud training. Tiny IoT devices (\eg, microcontrollers) typically have a limited SRAM size like 256KB. Such a small memory budget is hardly enough for the \emph{inference} of deep learning models~\cite{lin2020mcunet, lin2021mcunetv2, banbury2021micronets, burrello2021dory, liberis2020mu, fedorov2019sparse, liberis2019neural, rusci2019memory}, let alone the \emph{training}, which requires extra computation for the backward and extra memory for intermediate activation~\cite{chen2016training}. On the other hand, modern deep training frameworks (\eg, PyTorch~\cite{pytorch2019}, TensorFlow~\cite{tensorflow2015}) are usually designed for cloud servers and require a large memory footprint (>300MB) even when training a small model (\eg, MobileNetV2-w0.35~\cite{sandler2018mobilenetv2}) with batch size 1 (Figure.~\ref{fig:teaser}).

The huge gap (>1000$\times$) makes it impossible to run on tiny IoT devices with current frameworks and algorithms.  
Current deep learning training systems like PyTorch~\cite{pytorch2019}, TensorFlow~\cite{tensorflow2015}, JAX~\cite{jax2018github}, MXNet~\cite{chen2015mxnet}, \etc 
do not consider the tight resources on edge devices. Edge deep learning inference frameworks like TVM~\cite{chen2018tvm}, TF-Lite~\cite{tflite}, NCNN~\cite{tensorRT}, \etc provide a slim runtime, but lack the support for back-propagation.
Though there are low-cost efficient transfer learning algorithms like training only the final classifier layer, bias-only update~\cite{cai2020tinytl}, \etc, the accuracy drop is significant (Figure~\ref{fig:acc_vs_mem}), and existing training system can not realize the theoretical saving into measured saving. 
Furthermore, devices like microcontrollers are bare-metal and do not have an operational system and the runtime support needed by existing training frameworks. Therefore, we need to \textbf{jointly} design the \emph{algorithm} and the \emph{system} to enable tiny on-device training.

In this paper, we aim to bridge the gap and enable tiny on-device training with algorithm-system co-design. We investigate tiny on-device training and find two unique challenges: (1) the model is quantized on edge devices. A \emph{real} quantized graph is difficult to optimize due to low-precision tensors and the lack of Batch Normalization layers~\cite{ioffe2015batch}; (2) the limited hardware resource (memory and computation) of tiny hardware does not allow full back-propagation, whose memory usage can easily exceed the SRAM of microcontrollers by more than an order of magnitude. Only updating the last layer leads to poor accuracy (Figure~\ref{fig:acc_vs_mem}). To cope with the optimization difficulty, we propose \textit{\textbf{Quantization-Aware Scaling (QAS)}} to automatically scale the gradient of tensors with different bit-precisions, which effectively stabilizes the training and matches the accuracy of the floating-point counterpart (Section~\ref{sec:opt_mixed_prec}). QAS is hyper-parameter free and no tuning is required. To reduce the memory footprint of the full backward computation, we propose \textit{\textbf{Sparse Update}} to skip the gradient computation of less important layers and sub-tensors. We developed an automated method based on contribution analysis to find the best update scheme under different memory budgets (Section~\ref{sec:method:sparse_update}). Finally, we propose a lightweight training system, \textit{\textbf{\engine (\engineshort)}} , to implement the algorithm innovation (Section~\ref{sec:engine}). \engineshort is based on code generation; it offloads the auto-differentiation to the compile-time to greatly cut down the runtime overhead. It also supports advanced graph optimization like graph pruning and reordering to support sparse updates, achieving measured memory saving and speedup.

Our framework is the first solution to enable tiny on-device training of convolutional neural networks under 256KB SRAM and 1MB Flash without auxiliary memory. \textbf{(1)} Our solution enables weight update not only for the \emph{classifier} but also for the \emph{backbone}, which provides a \emph{high transfer learning accuracy} (Figure~\ref{fig:acc_vs_mem}). For tinyML application VWW~\cite{chowdhery2019visual}, our on-device finetuned model matches the accuracy of cloud training+edge deployment, and surpasses the common requirement of tinyML (MLPerf Tiny~\cite{banbury2020benchmarking}) by 9\%.
\textbf{(2)} Our system-algorithm co-design scheme effectively \emph{reduces the memory footprint}. As shown in Figure~\ref{fig:teaser}, the proposed techniques greatly reduce the memory usage by more than 1000$\times$ compared to PyTorch and Tensorflow. 
\textbf{(3)} Our framework also greatly \emph{accelerates training}, reducing the per-iteration time by more than 20$\times$ compared to dense update and vanilla system design (Figure~\ref{fig:latency_peakmem_comparison}).
\textbf{(4)} We deployed our training system to a Cortex M7 microcontroller STM32F746 to demonstrate the feasibility, suggesting that tiny IoT devices can not only perform inference but also training to adapt to new data. Our study paves the way for  lifelong on-device learning and opens up new possibilities for privacy-preserving device personalization.

\section{Approach}
\begin{figure*}[t]
    \centering
     \includegraphics[width=0.93\textwidth]{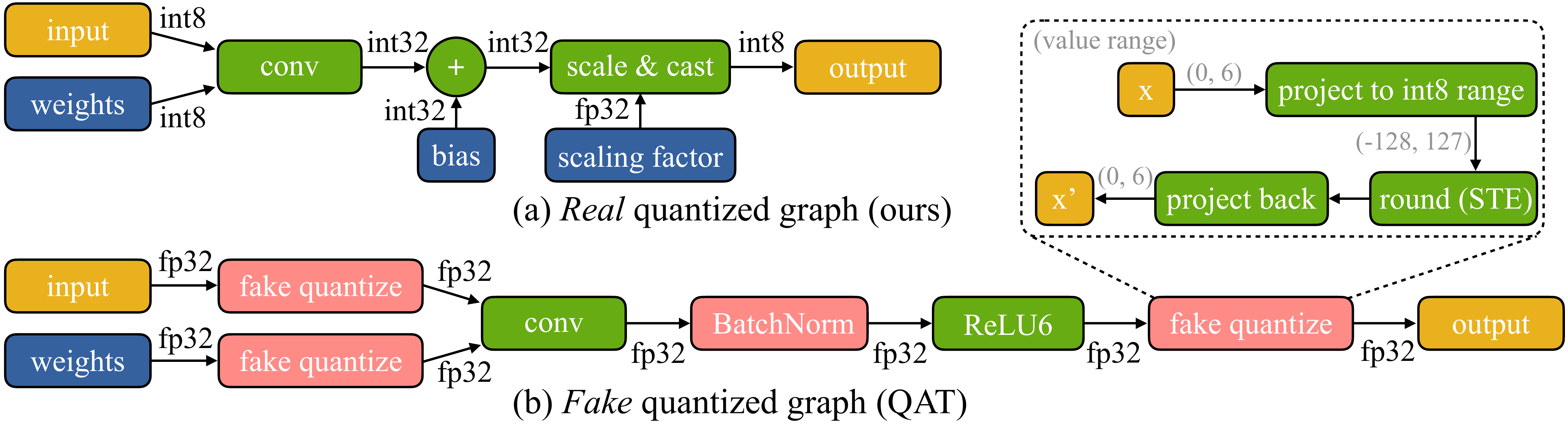}
    \caption{\emph{Real} quantized graphs (our optimized graph, designed for \emph{efficiency}) \vs \emph{fake} quantized graphs (for QAT, designed for \emph{simulation}). %
    The fake quantize graphs cannot provide memory saving due to floating-point operations. We need to use real quantized graph to fit the tight memory constraint. 
    }
    \label{fig:real_vs_fake_quant}
\end{figure*}

\myparagraph{Preliminaries.}
Neural networks usually need to be quantized to fit the limited memory of edge devices for inference~\cite{lin2020mcunet, jacob2018quantization}. 
For a \texttt{fp32} linear layer
$\mathbf{y}_{\texttt{fp32}} = \mathbf{W}_{\texttt{fp32}}\mathbf{x}_{\texttt{fp32}} + \mathbf{b}_{\texttt{fp32}}$, the \texttt{int8} quantized counterpart is:
\begin{equation}  \label{eq:quantize_forward}
\mathbf{\bar{y}}_{\texttt{int8}} = \texttt{cast2int8}[s_{\texttt{fp32}} \cdot(\mathbf{\bar{W}}_{\texttt{int8}}\mathbf{\bar{x}}_{\texttt{int8}} + \mathbf{\bar{b}}_{\texttt{int32}})],
\end{equation}\useshortskip
where $\bar{\cdot}$ denotes the tensor being quantized to fixed-point numbers, and $s$ is a floating-point scaling factor to project the results back into \texttt{int8} range. We call it \emph{real} quantized graphs (Figure~\ref{fig:real_vs_fake_quant}(a)) since tensors are in \texttt{int8} format.  
To keep the memory efficiency, we deploy and update the \emph{real} quantized graph on microcontrollers, and keep the updated weights as \texttt{int8}. The update formula is:
$\mathbf{\bar{W}^\prime}_{\texttt{int8}} = \texttt{cast2int8}(\mathbf{\bar{W}_{\texttt{int8}}} - \alpha\cdot \mathbf{G_\mathbf{\bar{W}}})$,
where $\alpha$ is the learning rate, and $\mathbf{G_\mathbf{\bar{W}}}$ is the gradient of the weights.
The gradient computation is also performed in \texttt{int8} for better computation efficiency. 

We update the real quantized graph for training, which is fundamentally different to quantization-aware training (QAT), where a \emph{fake} quantized graph (Figure~\ref{fig:real_vs_fake_quant}(b)) is trained on the cloud, and converted to a real one for deployment. As shown in Figure~\ref{fig:real_vs_fake_quant}(b), the fake quantization graph uses \texttt{fp32}, leading to no memory or computation savings. \emph{Real} quantized graphs are for \emph{efficiency}, while \emph{fake} quantized graphs are for \emph{simulation}. 
\label{sec:real_vs_fake_quantize}

\subsection{Optimizing Real Quantized Graphs}
\label{sec:opt_mixed_prec}
Unlike fine-tuning floating-point model on the cloud, training with \textit{a real} quantized graph
is difficult: the quantized graph has tensors of different bit-precisions (\texttt{int8}, \texttt{int32}, \texttt{fp32}, shown in Equation~\ref{eq:quantize_forward})
and lacks Batch Normalization~\cite{ioffe2015batch} layers (fused), leading to unstable gradient update.

\begin{figure*}[t]
    \centering
     \includegraphics[width=0.95\textwidth]{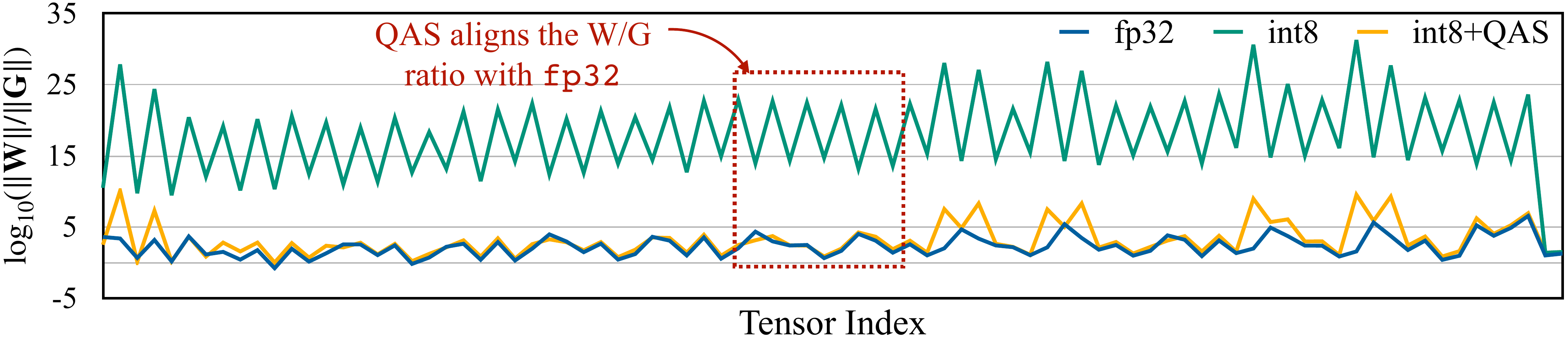}
    \caption{The quantized model has a very different weight/gradient norm ratio (\ie, $\lVert\mathbf{W}\rVert/\lVert\mathbf{G}\rVert$) compared to the floating-point model at training time. QAS stabilizes the $\lVert\mathbf{W}\rVert/\lVert\mathbf{G}\rVert$  ratio and helps optimization. For example, in the highlighted area, the ratios of the quantized model fluctuate dramatically in a zigzag pattern (weight, bias, weight, bias, ...); after applying QAS, the pattern stabilizes and matches the \texttt{fp32} counterpart.
    }
    \label{fig:per_layer_wg_rate}
\end{figure*}

\myparagraph{Gradient scale mismatch.}
When optimizing a quantized graph, the accuracy is lower compared to the floating-point counterpart. We hypothesize that the quantization process distorts the gradient update. 
To verify the idea, we plot the ratio between weight norm and gradient norm (\ie, $\lVert \mathbf{W}\rVert/\lVert\mathbf{G}\rVert$) for each tensor at the beginning of the training on the CIFAR dataset~\cite{krizhevsky2009learning} in Figure~\ref{fig:per_layer_wg_rate}. The ratio curve is very different after quantization: (1) the ratio is much larger (could be addressed by adjusting the learning rate); (2) the ratio has a different pattern after quantization. Take the highlighted area (red box) as an example, the quantized ratios have a zigzag pattern, differing from the floating-point curve.
If we use a fixed learning rate for all the tensors, then the update speed of each tensor would be very different compared to the floating-point case, leading to inferior accuracy. We empirically find that adaptive-learning rate optimizers like Adam~\cite{kingma2014adam} cannot fully address the issue (Section~\ref{sec:exp_qas}).

\myparagraph{Quantization-aware scaling (QAS).} 
To address the problem, we propose a hyper-parameter-free learning rate scaling rule, QAS.
Consider a 2D weight matrix of a linear layer $\mathbf{W}\in \mathds{R}^{c_1\times c_2}$, where $c_1, c_2$ are the input and output channel. To perform per-tensor quantization\footnote{For simplicity. We actually used per-channel quantization~\cite{jacob2018quantization} and the scaling factor is a vector of size $c_2$.}, we compute a scaling rate $s_\mathbf{W}\in\mathds{R}$, such that $\mathbf{\bar{W}}$'s largest magnitude is $2^7-1=127$: %
\begin{equation}
\mathbf{W} = s_\mathbf{W}\cdot(\mathbf{W} / s_\mathbf{W}) \stackrel{\text{quantize}}{\approx} s_\mathbf{W} \cdot \mathbf{\bar{W}}, \quad \mathbf{G_{\bar{W}}} \approx s_\mathbf{W}\cdot \mathbf{G_W},
\end{equation}
The process (roughly) preserves the mathematical functionality during the forward (Equation~\ref{eq:quantize_forward}), but it distorts the magnitude ratio between the weight and its corresponding gradient:
\begin{equation}\label{eq:wg_ratio}
  \lVert \mathbf{\bar{W}} \rVert / \lVert \mathbf{G_{\bar{W}}} \rVert \approx \lVert \mathbf{W}  /s_\mathbf{W} \rVert / \lVert s_\mathbf{W}\cdot \mathbf{G_W}\rVert = s_\mathbf{W}^{-2} \cdot \lVert\mathbf{W}\rVert / \lVert\mathbf{G}\rVert.
\end{equation}
We find that the weight and gradient ratios are off by  $s_\mathbf{W}^{-2}$, leading to the distorted pattern in Figure~\ref{fig:per_layer_wg_rate}: (1) the scaling factor is far smaller than 1, making the weight-gradient ratio much larger; (2) weights and biases have different data type (\texttt{int8} \vs \texttt{int32}) and thus have scaling factors of very different magnitude, leading to the zigzag pattern.
To solve the issue, we propose Quantization-Aware Scaling (QAS) by compensating the gradient of the quantized graph according to Equation~\ref{eq:wg_ratio}:
\begin{equation} \label{eq:qas}
    \mathbf{\Tilde{G}_{\bar{W}}} = \mathbf{G_{\bar{W}}} \cdot s_\mathbf{W}^{-2}, \quad 
    \mathbf{\Tilde{G}_{\bar{b}}} = \mathbf{G_{\bar{b}}} \cdot s_\mathbf{W}^{-2} \cdot s_{\mathbf{x}}^{-2} = \mathbf{G_{\bar{b}}} \cdot s^{-2}
\end{equation}
where $s_{\mathbf{X}}^{-2}$ is the scaling factor for quantizing input $\mathbf{x}$ (a scalar following~\cite{jacob2018quantization}, note that $s=s_\mathbf{W} \cdot s_{\mathbf{x}}$ in Equation~\ref{eq:quantize_forward}).
We plot the $\lVert \mathbf{W}\rVert/\lVert\mathbf{G}\rVert$ curve with QAS in Figure~\ref{fig:per_layer_wg_rate} (int8+scale). After scaling, the gradient ratios  match the floating-point counterpart. QAS enables fully quantized training (\texttt{int8} for both forward and backward) while matching  the accuracy of the floating-point training 
(Table~\ref{tab:optimizer_study}).

\subsection{Memory-Efficient Sparse Update} 
\label{sec:method:sparse_update}

\begin{figure*}[t]
    \centering
     \includegraphics[width=0.95\textwidth]{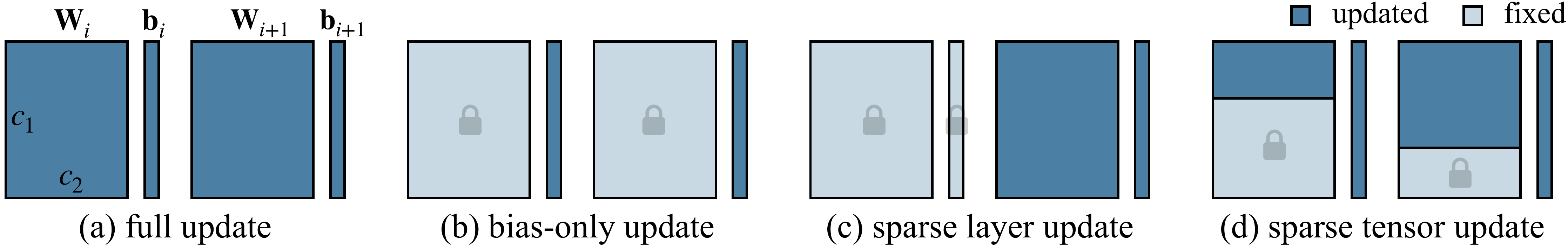}
     \vspace{-5pt}
    \caption{Different update paradigms of two linear layers in a deep neural network.
    }
    \label{fig:sparse_update}
\end{figure*}

Though QAS makes optimizing a quantized model possible, 
updating the whole model (or even the last several blocks) requires a large amount of memory, which is not affordable for the tinyML setting.  We propose to sparsely update the layers and the tensors.

\myparagraph{Sparse layer/tensor update.}
Pruning techniques prove to be quite successful for achieving sparsity and reducing model size~\cite{han2016deep, he2018amc, lin2017runtime, he2017channel, liu2017learning, liu2019metapruning}. 
Instead of pruning \emph{weights} for inference, we "prune" the \emph{gradient} during backpropagation, and update the model sparsely. Given a tight memory budget, we skip the update of the \emph{less important} parameters to reduce memory usage and computation cost. 
We consider updating a linear layer $\mathbf{y}=\mathbf{W}\mathbf{x}+\mathbf{b}$ (similar analysis applies to convolutions).
Given the output gradient $\mathbf{G_y}$ from the later layer, we can compute the gradient update by $\mathbf{G_W}=f_1(\mathbf{G_y}, \mathbf{x})$ and $\mathbf{G_b}=f_2(\mathbf{G_y})$. Notice that updating the biases does not require saving the intermediate activation $\mathbf{x}$, leading to a lighter memory footprint~\cite{cai2020tinytl}\footnote{If we update many layers, the intermediate activation could consume a large memory~\cite{chen2016training}.}; while updating the weights is more memory-intensive but also more expressive. For hardware like microcontrollers, we also need an extra copy for the updated parameters since the original ones are stored in read-only FLASH~\cite{lin2020mcunet}.
Given the different natures of updating rules, we consider the sparse update rule in three aspects (Figure~\ref{fig:sparse_update}):
(1) \emph{Bias update}: how many layers should we backpropagate to and update the biases (bias update is cheap, we always update the biases if we have backpropagated to a layer).
(2) \emph{Sparse layer update}: select a subset of layers to update the corresponding weights.
(3) \emph{Sparse tensor update}: we further allow updating a subset of weight channels to reduce the cost. %

However, finding the right sparse update scheme under a memory budget is challenging due to the large combinational space. For MCUNet~\cite{lin2020mcunet} model with 43 convolutional layers and weight update ratios from \{0, 1/8, 1/4, 1/2, 1\},
the combination is about $10^{30}$, making exhaustive search impossible.

\begin{figure*}[t]
    \centering
     \includegraphics[width=\textwidth]{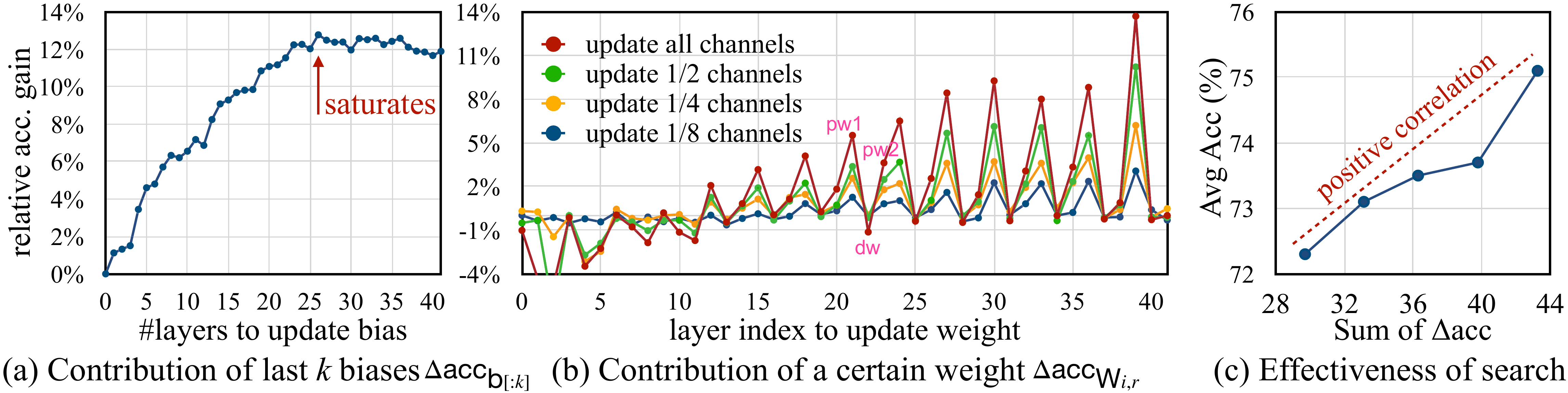}
    \caption{Contribution analysis of updating biases and weights. (a) For bias update, the accuracy generally goes higher as more layers are updated, but plateaus soon. (b) For updating the weight of a specific layer, the later layers appear to be more important; the first point-wise conv (pw1) in an inverted bottleneck block~\cite{sandler2018mobilenetv2} appears to be more important; and the gains are bigger with more channels updated.
    (c) The automated selection based on contribution analysis is effective: the actual downstream accuracy shows a positive correlation with  $\sum\Delta\text{acc}$. %
    }
    \label{fig:sensitivity_curve}
\end{figure*}

\myparagraph{Automated selection with contribution analysis.} 
We propose to automatically derive the sparse update scheme by \emph{contribution analysis}. We find the contribution of each parameter (weight/bias) to the downstream accuracy.
Given a convolutional neural network with $l$ layers, we measure the accuracy improvement from (1) biases: the improvement of updating \emph{last} $k$ biases $\textbf{b}_{l}, \textbf{b}_{l-1},..., \textbf{b}_{l-k+1}$ (bias-only update) compared to only updating the classifier, defined as $\Delta\text{acc}_{\textbf{b}[:k]}$; (2) weights: the improvement of updating the weight of one extra layer $\textbf{W}_i$ (with a channel update ratio $r$) compared to bias-only update, defined as $\Delta\text{acc}_{\textbf{W}i, r}$. An example of the contribution analysis can be found in Figure~\ref{fig:sensitivity_curve} (MCUNet on Cars~\cite{krause20133d} dataset; please find more results in appendix Section~\ref{sec:more_contribution}).
After we find $\Delta\text{acc}_{\textbf{b}[:k]}$ and $\Delta\text{acc}_{\textbf{W}i}$ ($1\leq k, i\leq l$), we solve an optimization problem to find:
\begin{equation}
    k^*, \mathbf{i}^*, \mathbf{r}^* = \max_{k, \mathbf{i}, \mathbf{r}} (\Delta\text{acc}_{\textbf{b}[:k]} + \sum_{i\in\mathbf{i}, r\in\mathbf{r}} \Delta\text{acc}_{\textbf{W}i, r}) \quad
    \text{s.t. Memory}(k, \mathbf{i}, \mathbf{r}) \leq \text{constraint},
\end{equation}
where $\mathbf{i}$ is a collection of layer indices whose weights are updated, and $\mathbf{r}$ is the corresponding update ratios (1/8, 1/4, 1/2, 1). Intuitively, by solving this optimization problem, we find the combination of (\#layers for bias update, the subset of weights to update), such that the total contribution are maximized  while the memory overhead does not exceed the constraint. The problem can be efficiently solved with evolutionary search (see Section~\ref{sec:evo_vs_random}). 
Here we assume that the accuracy contribution of each tensor ($\Delta\text{acc}$) can be summed up. Such approximation is quite effective (Figure~\ref{fig:sensitivity_curve}(c)).

\subsection{\engine (\engineshort)}
\label{sec:engine}

\begin{figure*}[t]
    \centering
     \includegraphics[width=1\textwidth]{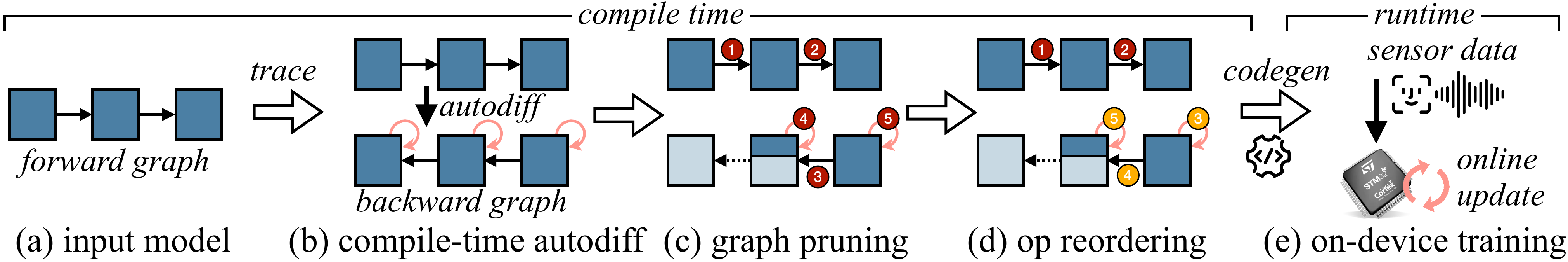}
    \caption{
    The workflow of our \engine (\engineshort). 
    \textbf{(a,b)} Our engine traces the forward graph for a given model and derives the corresponding backward graph at compile time. The {\color{fig5red} red} cycles denote the gradient descent operators.
    \textbf{(c)} To reduce memory requirements, nodes related with frozen weights (colored in {\color{fig5gray} light blue}) are pruned from backward computation.
    \textbf{(d)} To minimize memory footprint, the gradient descent operators are re-ordered to be interlaced with backward computations (colored in {\color{fig5yellow} yellow}).
    \textbf{(e)} \engineshort compiles forward and backward graphs using code generation and deploys training on tiny IoT devices (best viewed in colors).
 }
    \label{fig:compiler_stack}
\end{figure*}

The theoretical saving from real quantized training and sparse update does not translate to measured memory saving in existing deep learning frameworks, due to the redundant runtime and the lack of graph pruning.
We co-designed an efficient training system, \engine (\engineshort), to transform the above algorithms into slim binary codes (Figure~\ref{fig:compiler_stack}). %

\myparagraph{Compile-time differentiation and code generation.}
\engineshort offloads the auto-differentiation from the runtime to the compile-time, generating a static backward graph which can be pruned and optimized (see below) to reduce the memory and computation. \engineshort is based on code generation: it compiles the optimized graphs to executable binaries on the target hardware, which minimizes the runtime library size and removes the need for host languages like Python (typically uses Megabytes of memory).

\myparagraph{Backward graph pruning for sparse update.}
We prune the redundant nodes  in the backward graph before compiling it to binary codes.
For sparse layer update, we prune away the gradient nodes of the frozen weights, only keeping the nodes for bias update. Afterwards, we traverse the graph to find unused intermediate nodes due to pruning (\eg, saved input activation) and apply dead-code elimination (DCE) to remove the redundancy. 
For sparse tensor update, we introduce a \emph{sub-operator slicing} mechanism to split a layer's weights into trainable and frozen parts; the backward graph of the frozen subset is removed. 
Our compiler translates the sparse update algorithm into measured memory saving, reducing the training memory 7-9$\times$ without losing accuracy (Figure~\ref{fig:latency_peakmem_comparison}(a), blue v.s. yellow).

\begin{figure*}[t]
    \centering
     \includegraphics[width=\textwidth]{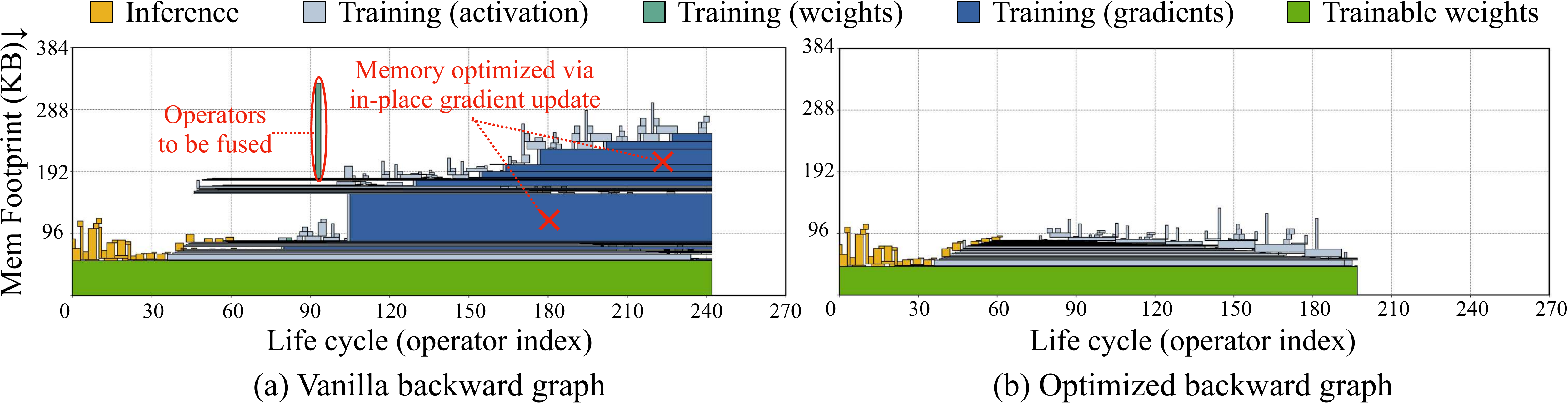}
    \caption{Memory footprint reduction by operator reordering. 
    With operator reordering, \engineshort can apply in-place gradient update and perform operator fusion to avoid large intermediate tensors to reduce memory footprint. We profiled MobileNetV2-w0.35 in this figure (same as Figure~\ref{fig:teaser}).
     }
    \label{fig:mem_footprint}
\end{figure*}
\myparagraph{Operator reordering and in-place update.}
The execution order of different operations affects the life cycle of tensors and the overall memory footprint. This has been well-studied for inference~\cite{ahn2020ordering, liberis2019neural} but not for training due to the extra complexity.
Traditional training frameworks usually derive the gradients of all the trainable parameters before applying the update. Such a practice leads to significant memory waste for storing the gradients. By reordering operators, we can immediately apply the gradient update to a specific tensor (in-place update) before back-propagating to earlier layers, so that the gradient can be released. As such, we trace the dependency of all tensors (weights, gradients, activation) and reorder the operators, so that some operators can be fused to reduce memory footprint (by 2.4-3.2$\times$, Figure~\ref{fig:latency_peakmem_comparison}(a), yellow v.s. red). The memory life cycle analysis in Figure~\ref{fig:mem_footprint} reflects the memory saving from in-place gradient update and operator fusion.

\vspace{-5pt}
\section{Experiments}

\subsection{Setups}
\label{sec:exp_setup}
\myparagraph{Training.}
We used three popular tinyML models in our experiments: MobileNetV2~\cite{sandler2018mobilenetv2} (width multiplier 0.35, backbone 17M MACs, 0.25M Param), 
ProxylessNAS~\cite{cai2019proxylessnas} (width multiplier 0.3, backbone 19M MACs, 0.33M Param), 
MCUNet~\cite{lin2020mcunet} (the 5FPS ImageNet model, backbone 23M MACs, 0.48M Param).
We pre-trained the models on ImageNet~\cite{deng2009imagenet} and perform post-training quantization~\cite{jacob2018quantization}. The quantized models are fine-tuned on downstream datasets to evaluate the transfer learning capacity. 
We perform the training and memory/latency measurement on a microcontroller STM32F746 (320KB SRAM, 1MB Flash) using a single batch size. To faster obtain the accuracy statistics on multiple downstream datasets, we simulate the training results on GPUs, and we verified that the simulation obtains the same level of accuracy compared to training on microcontrollers.
Please refer to the the appendix (Section~\ref{sec:training_details}) for detailed training hyper-parameters.
We also provide a \emph{video demo} of deploying our training system on microcontroller in the appendix (Section~\ref{sec:video_demo}).

\myparagraph{Datasets.}
We measure the transfer learning accuracy on multiple downstream datasets and report the average accuracy~\cite{kolesnikov2020big}. We follow~\cite{cai2020tinytl} to use a set of vision datasets including Cars~\cite{krause20133d}, CIFAR-10~\cite{krizhevsky2009learning}, CIFAR-100~\cite{krizhevsky2009learning}, CUB~\cite{cub}, Flowers~\cite{nilsback2008automated}, Food~\cite{bossard2014food}, and Pets~\cite{parkhi2012cats}\footnote{Pets uses \href{https://creativecommons.org/licenses/by-sa/4.0/}{CC BY-SA 4.0} license; Cars and ImageNet use \href{https://image-net.org/download.php}{the ImageNet license}; others are not listed.}. We fine-tuned the models
on all these datasets for 50 epochs following~\cite{cai2020tinytl}. We also include VWW dataset~\cite{chowdhery2019visual}, a widely used benchmark for tinyML applications. We train on VWW for 10 epochs following~\cite{lin2020mcunet}. We used resolution 128 for all datasets and models for a fair comparison.

\myparagraph{Memory estimation.}
The memory usage of a computation graph is related to its implementation~\cite{ahn2020ordering, liberis2019neural, lin2020mcunet, lin2021mcunetv2}. We provide two settings for memory measurement: (1) \textbf{analytic profiling}: we count the size of \emph{extra} tensors required for backward computation, including the saved intermediate activation, binary truncation task, and the updated weights. The size is implementation-agnostic. It is used for a fast profiling;  (2) \textbf{on-device profiling}: we measure the actual memory usage when running model training on an STM32F746 MCU (320KB SRAM, 1MB Flash). We used TinyEngineV2~\cite{lin2021mcunetv2} as the backend and 2$\times$2 patch-based inference~\cite{lin2021mcunetv2} for the initial stage to reduce the forward peak memory. The \emph{measured} memory determines whether a solution can be deployed on the hardware.

\subsection{Experimental Results}
\label{sec:exp_qas}

\begin{figure*}[t]
    \centering
     \includegraphics[width=0.85\textwidth]{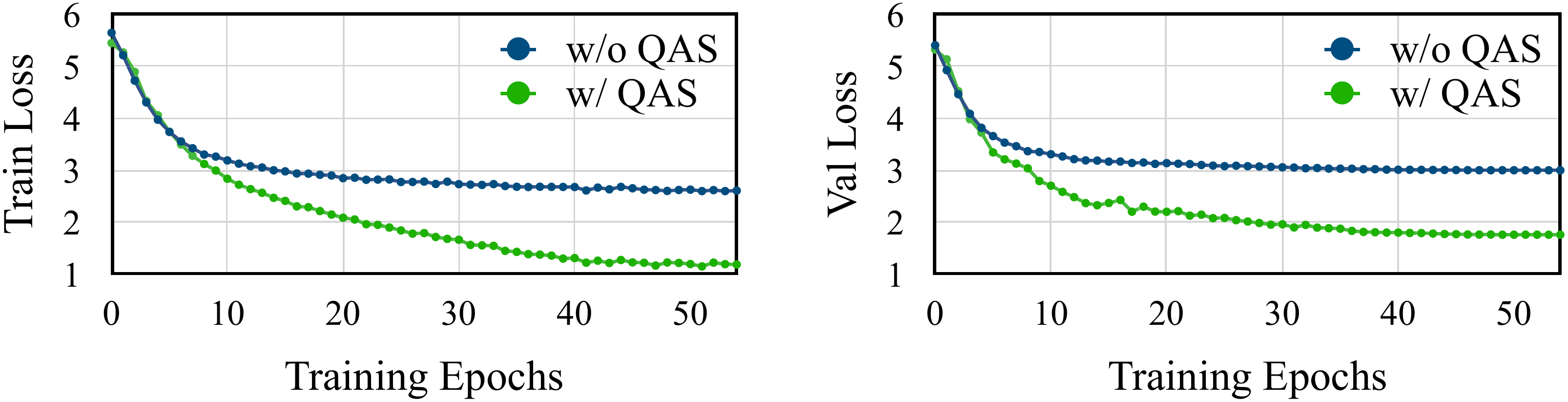}
    \caption{Training and validation loss curves w/ and w/o QAS. QAS effectively helps convergence, leading to better accuracy. The results are from updating the last two blocks of the MCUNet model on the Cars dataset.
    }
    \label{fig:training_curves_short}
\end{figure*}

\myparagraph{Quantization-aware scaling (QAS) addresses the optimization difficulty.}
We fine-tuned the last two blocks (simulate low-cost fine-tuning) of MCUNet to various downstream datasets (Table~\ref{tab:optimizer_study}). With momentum SGD, the training accuracy of the quantized model (\texttt{int8}) falls behind the floating-point counterpart due to the optimization difficulty.
Adaptive learning rate optimizers like Adam~\cite{kingma2014adam}  can improve the accuracy but are still lower than the \texttt{fp32} fine-tuning results; it also costs \textbf{3$\mathbf{\times}$ memory} consumption due to second-order momentum, which is not desired for tinyML settings. LARS~\cite{you2017large} cannot converge well on most datasets despite extensive hyper-parameter tuning (over both learning rate and the "trust coefficient"). We hypothesize that the aggressive gradient scaling rule of LARS makes the training unstable.
The accuracy gap is closed when we apply QAS, matching the accuracy of floating-point training at no extra memory cost.
The learning curves (fine-tuning) of MCUNet on the Cars dataset w/ and w/o QAS are also provided in Figure~\ref{fig:training_curves_short}. Therefore, QAS effectively helps optimization.

\begin{table}[t]
    \caption{Updating real quantized graphs (\texttt{int8}) for the fine-tuning is difficult: the accuracy falls behind the floating-point counterpart (\texttt{fp32}), even with adaptive learning rate optimizers like Adam~\cite{kingma2014adam} and LARS~\cite{you2017large}. QAS helps to bridge the accuracy gap without memory overhead (slightly higher due to randomness). The numbers are for updating the last two blocks of MCUNet-5FPS~\cite{lin2020mcunet} model.
    }
    \label{tab:optimizer_study}
    \centering
    \small{
     \begin{tabular}{llccccccccc}
    \toprule
 \multirow{2}{*}{Precision} & \multirow{2}{*}{Optimizer} &  \multicolumn{8}{c}{Accuracy (\%) (MCUNet backbone: 23M MACs, 0.48M Param )} & \multirow{2}{*}{\shortstack{Avg\\Acc.}} \\ \cmidrule(lr){3-10}
 & & Cars & CF10 & CF100 & CUB & Flowers & Food & Pets & VWW \\
\midrule
\texttt{fp32} & SGD-M & 56.7 & 86.0 & 63.4 & 56.2 & 88.8 & 67.1 & 79.5 & 88.7 & 73.3 \\ \midrule 
 \multirow{4}{*}{\texttt{int8}} & SGD-M &  31.2 & 75.4 & 54.5 & 55.1 & 84.5 & 52.5 & 81.0 & 85.4 & 64.9 \\ %
 & Adam~\cite{kingma2014adam} &  54.0 & 84.5 & 61.0 & 58.5 & 87.2 & 62.6 & 80.1 & 86.5 & 71.8  \\
 & LARS~\cite{you2017large} & 5.1 & 64.8 & 39.5 & 9.6 & 28.8 & 46.5 & 39.1 & 85.0 & 39.8  \\  \cmidrule(lr){2-11}
& SGD-M+QAS & 55.2 & 86.9 & 64.6 & 57.8 & 89.1 & 64.4 & 80.9 & 89.3 & \textbf{73.5} \\
    \bottomrule
     \end{tabular}
     }
\end{table}

\begin{figure*}[t]
    \centering
     \includegraphics[width=\textwidth]{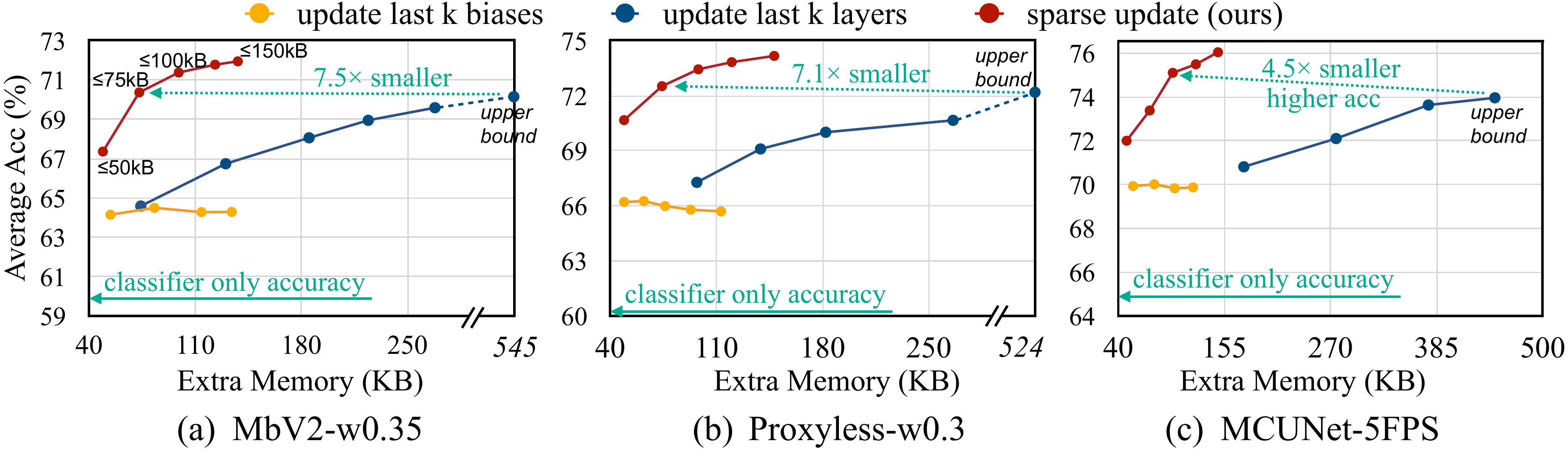}
    \caption{Sparse update can achieve higher transfer learning accuracy using 4.5-7.5$\times$ smaller extra memory (analytic) compared to updating the last $k$ layers. For classifier-only update, the accuracy is low due to limited capacity. Bias-only update can achieve a higher accuracy but plateaus soon. 
    }
    \label{fig:acc_vs_mem}
\end{figure*}
\myparagraph{Sparse update obtains better accuracy at lower memory.}
We compare the performance of our searched sparse update schemes with two baseline methods: fine-tuning only biases of the last $k$ layers; fine-tuning weights and biases of the last $k$ layers (including fine-tuning the full model, when $k$ equals to the total \#layers). For each configuration, we measure the average accuracy on the 8 downstream datasets and the \emph{analytic} extra memory usage. We also compare with a simple baseline by only fine-tuning the classifier. As shown in Figure~\ref{fig:acc_vs_mem}, the accuracy of classifier-only update is low due to the limited learning capacity. \emph{Updating the classifier alone is not enough; we also need to update the backbone.} Bias-only update outperforms classifier-only update but the accuracy quickly plateaus and does not improve even more biases are tuned. For updating last $k$ layers, the accuracy generally goes higher as more layers are tuned; however, it has a very large memory footprint. Take MCUNet as an example, updating the last two blocks leads to an extra memory surpassing 256KB, making it infeasible for IoT devices/microcontrollers.
Our sparse update scheme can achieve higher downstream accuracy at a much lower memory cost: compared to updating last $k$ layers, sparse update can achieve higher downstream accuracy with smaller memory footprint. 
We also measure the highest accuracy achievable by updating the last $k$ layers (including fine-tuning the full model\footnote{Note that fine-tuning the entire model does not always lead to the best accuracy. We grid search for the best $k$ on Cars dataset: $k=$36 for MobileNetV2, 39 for ProxylessNAS, 12 for MCUNet, and apply it to all datasets.}) as the baseline upper bound (denoted as "upper bound"). Interestingly, our sparse update achieves a better downstream accuracy compared to the baseline best statistics. We hypothesize that the sparse update scheme alleviates over-fitting or makes momentum-free optimization easier. 

\myparagraph{Matching cloud training accuracy for tinyML.} Remarkably, the downstream accuracy of our on-device training has \emph{matched or even surpassed} the accuracy of cloud-trained results on tinyML application VWW~\cite{chowdhery2019visual}. Our framework uses 206KB \emph{measured} SRAM while achieving 89.1\% top-1 accuracy for on-device training (we used gradient accumulation for the VWW dataset; see the appendix Section~\ref{sec:training_details} for details). The result is higher than the accuracy of the same model reported by the state-of-the-art solution MCUNet (88.7\%, trained on cloud and deployed to MCU). Both settings transfer the ImageNet pre-trained model to VWW. The on-device accuracy is far above the common requirement for tinyML (>80\% by MLPerf Tiny~\cite{banbury2020benchmarking}) and surpassed the results of industry solution TF-Lite Micro+MobileNetV2 (86.2\%~\cite{lin2020mcunet} under 256KB, \emph{inference-only, no  training support}).

\begin{figure*}[t]
    \centering
     \includegraphics[width=\textwidth]{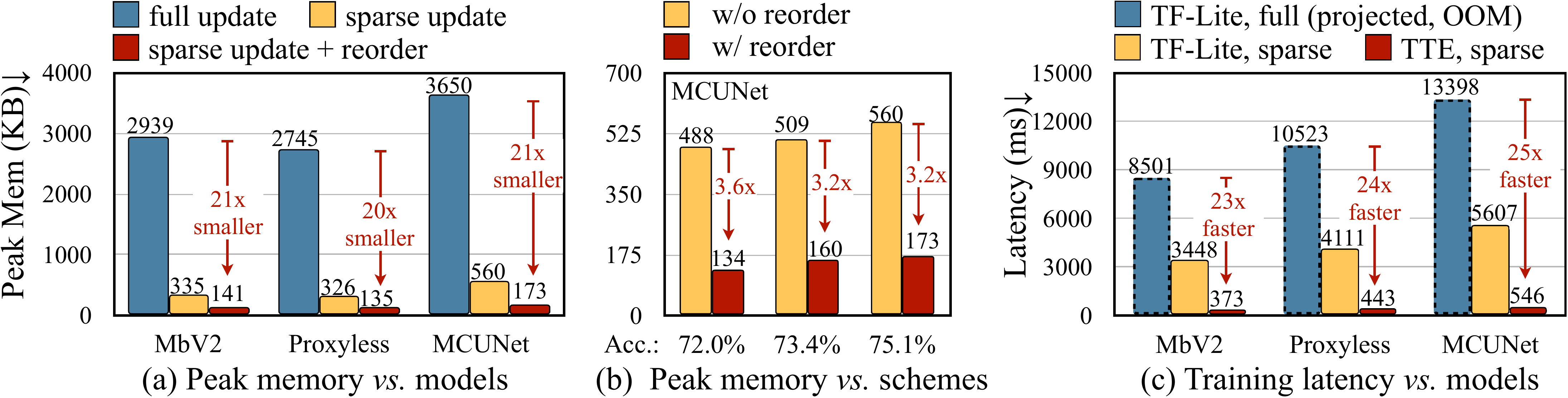}
    \caption{\emph{Measured} peak memory and latency: \textbf{(a)} Sparse update with TTE graph optimization can reduce the measured peak memory by 20-21$\times$ for different models, making training feasible on tiny edge devices.
    \textbf{(b)} Graph optimization consistently reduces the peak memory for different sparse update schemes (denoted by different average transfer learning accuracies).
    \textbf{(c)} Sparse update with TTE operators achieves 23-25$\times$ faster training speed compared to the full update with TF-Lite Micro operators, leading to less energy usage.
    \emph{Note}: for sparse update, we choose the config that achieves the same accuracy as full update.
    }
    \label{fig:latency_peakmem_comparison}
\end{figure*}
\myparagraph{\engine: memory saving.}
We measure the training memory of three models on STM32F746 MCU to compare the memory saving from \engineshort. We measure the peak SRAM usage under three settings: general full update, sparse update, and sparse update with TTE graph reordering (Figure~\ref{fig:latency_peakmem_comparison}(a)).
The sparse update effectively reduces peak memory by 7-9$\times$ compared to the full update thanks to the graph pruning mechanism, while achieving the same or higher transfer learning accuracy (compare the data points connected by arrows in Figure~\ref{fig:acc_vs_mem}). The memory is further reduced with operator reordering, leading to 20-21$\times$ total memory saving. With both techniques, the training of all 3 models fits 256KB SRAM.
We also compare the memory saving of reordering under different update schemes on MCUNet (Figure~\ref{fig:acc_vs_mem}(b), indicated by different accuracy levels).
Reordering consistently reduces the peak memory for different sparse update schemes of varying learning capacities.

\myparagraph{\engine: faster training.}
We further measure the training latency per image on the STM32F746 MCU with three settings: full update with TF-Lite Micro kernels, sparse update with TF-Lite Micro kernels, and sparse update with TTE kernels (Figure~\ref{fig:latency_peakmem_comparison}(c)). Notice that TF-Lite \emph{does not} support training; we just used the kernel implementation to measure latency.
By graph optimization and exploiting multiple compiler optimization approaches (such as loop unrolling and tiling), our sparse update + TTE kernels can significantly enhance the training speed by 23-25$\times$ compared to the full update + TF-Lite Micro kernels, leading to energy saving and making training practical. Note that TF-Lite with full update leads to OOM, so we report the projected latency according to the average speed of each op type (marked in dashed columns).

\subsection{Ablation Studies and Analysis}

\begin{figure*}[t]
    \centering
     \includegraphics[width=\textwidth]{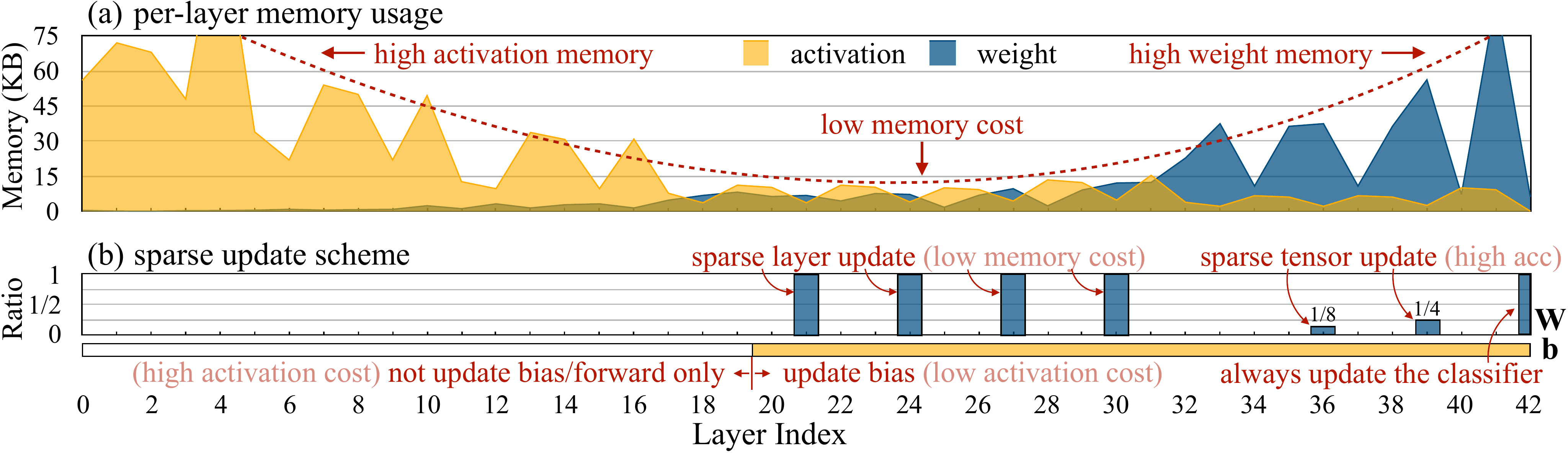}
    \caption{\textbf{(a)} The weight and activation memory cost of updating \emph{each} layer of MCUNet (analytic). We find that the activation cost is high for the starting layers; the weight cost is high for the later layers; the overall memory cost is low for the middle layers. \textbf{(b)} Dissecting the sparse update scheme: we update the biases of the last 22 layers due to its low activation cost. For weight update, we update some middle layers due to its low memory cost, and update partial channels of the two later layers since they are important for accuracy (Figure~\ref{fig:sensitivity_curve}).
    } 
    \label{fig:update_scheme}
\end{figure*}
\myparagraph{Dissecting update schedules.}
We visualize the update schedule of the MCUNet~\cite{lin2020mcunet} model searched under 100KB extra memory (analytic) in Figure~\ref{fig:update_scheme} (lower subfigure (b), with 10 classes). It updates the biases of the last 22 layers, and sparsely updates the weights of 6 layers (some are sub-tensor update).
The initial 20 layers are frozen and run forward only. To understand why this scheme makes sense, we also plot the memory cost from activation and weight when updating \emph{each} layer in the upper subfigure (a). We see a clear pattern: the activation cost is high for the initial layers; the weight cost is high for the ending layers; while the total memory cost is low when we update the middle layers (layer index 18-30). The update scheme matches the memory pattern: to skip the initial stage of high activation memory, we only update biases of the later stage of the network; we update the weights of 4 intermediate layers due to low overall memory cost; we also update the partial weights of two later layers (1/8 and 1/4 weights) due to their high contribution to the downstream accuracy (Figure~\ref{fig:sensitivity_curve}). Interestingly, all the updated weights are from the first point-wise convolution in each inverted residual block~\cite{sandler2018mobilenetv2} as they generally have a higher contribution to accuracy (the peak points on the zigzag curve in Figure~\ref{fig:sensitivity_curve}(b)).

\myparagraph{Effectiveness of contribution analysis.} We verify if the update scheme search based on contribution analysis is effective. We collect several data points during the search process (the update scheme and the search criteria, \ie, the sum of $\Delta\text{acc}$). We train the model with each update scheme to get the average accuracy on the downstream datasets (the real optimization target) and plot the comparison in Figure~\ref{fig:sensitivity_curve}(c). We observe a positive correlation, indicating the effectiveness of the search.

\myparagraph{Sub-channel selection.}
Similar to weight pruning, we need to select the subset of channels for sub-tensor update. 
We update the last two blocks of the MCUNet~\cite{lin2020mcunet} model and only 1/4 of the weights for each layer to compare the accuracy of different channel selection methods (larger magnitude, smaller magnitude, and random). The results are quite similar (within 0.2\% accuracy difference). Channel selection is not very important for transfer learning (unlike pruning). 
We choose to update the channels with a larger weight magnitude since it has slightly higher accuracy.

\section{Related Work}

\myparagraph{Efficient transfer learning.}
There are several ways to reduce the transfer learning cost compared to fine-tuning the full model~\cite{kornblith2019better,cui2018large, kolesnikov2020big}. 
The most straightforward way is to only update the classifier layer~\cite{chatfield2014return,donahue2014decaf,gan2015devnet,sharif2014cnn}, but the accuracy is low when the domain shift is large~\cite{cai2020tinytl}. 
Later studies investigate other tuning methods including updating biases~\cite{cai2020tinytl, zaken21bitfit}, updating normalization layer parameters~\cite{mudrakarta2019k,frankle2020training}, updating small parallel branches~\cite{cai2020tinytl, hu2021lora}, \etc. 
These methods only reduce the trainable parameter number but lack the study on system co-design to achieve real memory savings. Most of them do not fit tinyML settings (cannot handle quantized graph and lack of BatchNorm~\cite{ioffe2015batch}).

\myparagraph{Systems for deep learning.}
The success of deep learning is built on top of popular training frameworks such as  PyTorch~\cite{pytorch2019}, TensorFlow~\cite{abadi2016tensorflow}, MXNet~\cite{chen2015mxnet}, JAX~\cite{jax2018github}, \etc. These systems usually depend on a host language (\eg Python) and various runtime, which brings 
significant overhead (>300MB) and does not fit tiny edge devices.
Inference libraries like TVM~\cite{chen2018tvm}, TF-Lite~\cite{tflite}, NCNN~\cite{ncnn}, TensorRT~\cite{tensorRT}, and OpenVino~\cite{vaswani2017attention} provide lightweight runtime environments but do not support training; MNN~\cite{jiang2020mnn} has a preliminary support for full model training but cannot fit tiny IoT devices.
Recently, POET~\cite{patil2022poet} utilizes rematerialization and paging to train on microcontrollers, but it relies on a large external memory.

\myparagraph{Tiny deep learning on microcontrollers.}
Tiny deep learning on microcontrollers is challenging. 
Existing work explores model compression (pruning~\cite{han2016deep, he2018amc, lin2017runtime, he2017channel, liu2017learning, yu2017scalpel, liberis2021differentiable}, quantization~\cite{han2016deep, rastegari2016xnor, wang2019haq, choi2018pact, rusci2019memory, langroudi2021tent, lin2020mcunet, jacob2018quantization}) and neural architecture search~\cite{zoph2017neural, zoph2018learning, tan2019mnasnet, lin2020mcunet, banbury2021micronets, liberis2020mu, fedorov2019sparse, lyu2021resource, lin2020mcunet, lin2021mcunetv2} to reduce the required resource of deep learning models. There are several deep learning systems for tinyML (TF-Micro~\cite{abadi2016tensorflow}, CMSIS-NN~\cite{lai2018cmsis}, TinyEngine~\cite{lin2020mcunet}, MicroTVM~\cite{chen2018tvm}, CMix-NN~\cite{capotondi2020cmix}, \etc). However, the above algorithms and systems are only for inference but not training. 
There are several preliminary attempts to explore training on microcontrollers~\cite{ren2021tinyol, grau2021device, sudharsan2021train, sudharsan2021globe2train}. However, due to the lack of efficient algorithm and system support, they are only able to tune one layer or a very small model, while our work supports the tuning of modern CNNs for real-life applications.

\section{Conclusion}
\label{sec:conclusion}
In this paper, we propose the first solution to enable tiny on-device training on microcontrollers under a tight memory budget of 256KB and 1MB Flash without auxiliary memory. Our algorithm system co-design solution significantly reduces the training memory (more than 1000$\times$ compared with PyTorch and TensorFlow) and per-iteration latency (more than 20$\times$ speedup over TensorFlow-Lite Micro), allowing us to obtain higher downstream accuracy. Our study suggests that tiny IoT devices can not only perform inference but also continuously adapt to new data for lifelong learning.

\myparagraph{Limitations and societal impacts.}
Our work achieves the first practical solution for transfer learning on tiny microcontrollers. However, our current study is limited to vision recognition with CNNs. In the future, we would like to extend to more modalities (\eg, audio) and more models (\eg, RNNs, Transformers). 
Our study improves tiny on-device learning, which helps to protect the privacy on sensitive data (\eg, healthcare). However, to design and benchmark our method, we experimented on many downstream datasets, leading to a fair amount of electricity consumption. 

\section*{Acknowledgments}
We thank National Science Foundation (NSF), MIT-IBM Watson AI Lab, MIT AI Hardware Program, Amazon, Intel, Qualcomm, Ford, Google for supporting this research.

\small
\bibliographystyle{plain}
\bibliography{main}

\begin{thebibliography}{10}

\bibitem{ncnn}
Ncnn : A high-performance neural network inference computing framework
  optimized for mobile platforms.
\newblock \url{https://github.com/Tencent/ncnn}.

\bibitem{tensorRT}
Nvidia tensorrt, an sdk for high-performance deep learning inference.
\newblock \url{https://developer.nvidia.com/tensorrt}.

\bibitem{tflite}
Tensorflow lite.
\newblock \url{https://www.tensorflow.org/lite}.

\bibitem{tensorflow2015}
Mart\'{\i}n Abadi, Ashish Agarwal, Paul Barham, Eugene Brevdo, Zhifeng Chen,
  Craig Citro, Greg~S. Corrado, Andy Davis, Jeffrey Dean, Matthieu Devin,
  Sanjay Ghemawat, Ian Goodfellow, Andrew Harp, Geoffrey Irving, Michael Isard,
  Yangqing Jia, Rafal Jozefowicz, Lukasz Kaiser, Manjunath Kudlur, Josh
  Levenberg, Dandelion Man\'{e}, Rajat Monga, Sherry Moore, Derek Murray, Chris
  Olah, Mike Schuster, Jonathon Shlens, Benoit Steiner, Ilya Sutskever, Kunal
  Talwar, Paul Tucker, Vincent Vanhoucke, Vijay Vasudevan, Fernanda Vi\'{e}gas,
  Oriol Vinyals, Pete Warden, Martin Wattenberg, Martin Wicke, Yuan Yu, and
  Xiaoqiang Zheng.
\newblock {TensorFlow}: Large-scale machine learning on heterogeneous systems,
  2015.
\newblock Software available from tensorflow.org.

\bibitem{abadi2016tensorflow}
Mart{\'\i}n Abadi, Paul Barham, Jianmin Chen, Zhifeng Chen, Andy Davis, Jeffrey
  Dean, Matthieu Devin, Sanjay Ghemawat, Geoffrey Irving, Michael Isard, et~al.
\newblock Tensorflow: A system for large-scale machine learning.
\newblock In {\em OSDI}, 2016.

\bibitem{ahn2020ordering}
Byung~Hoon Ahn, Jinwon Lee, Jamie~Menjay Lin, Hsin-Pai Cheng, Jilei Hou, and
  Hadi Esmaeilzadeh.
\newblock Ordering chaos: Memory-aware scheduling of irregularly wired neural
  networks for edge devices.
\newblock {\em arXiv preprint arXiv:2003.02369}, 2020.

\bibitem{banbury2021micronets}
Colby Banbury, Chuteng Zhou, Igor Fedorov, Ramon Matas, Urmish Thakker, Dibakar
  Gope, Vijay Janapa~Reddi, Matthew Mattina, and Paul Whatmough.
\newblock Micronets: Neural network architectures for deploying tinyml
  applications on commodity microcontrollers.
\newblock {\em Proceedings of Machine Learning and Systems}, 3, 2021.

\bibitem{banbury2020benchmarking}
Colby~R Banbury, Vijay~Janapa Reddi, Max Lam, William Fu, Amin Fazel, Jeremy
  Holleman, Xinyuan Huang, Robert Hurtado, David Kanter, Anton Lokhmotov,
  et~al.
\newblock Benchmarking tinyml systems: Challenges and direction.
\newblock {\em arXiv preprint arXiv:2003.04821}, 2020.

\bibitem{bossard2014food}
Lukas Bossard, Matthieu Guillaumin, and Luc~Van Gool.
\newblock Food-101--mining discriminative components with random forests.
\newblock In {\em European conference on computer vision}, pages 446--461.
  Springer, 2014.

\bibitem{jax2018github}
James Bradbury, Roy Frostig, Peter Hawkins, Matthew~James Johnson, Chris Leary,
  Dougal Maclaurin, George Necula, Adam Paszke, Jake Vander{P}las, Skye
  Wanderman-{M}ilne, and Qiao Zhang.
\newblock {JAX}: composable transformations of {P}ython+{N}um{P}y programs,
  2018.

\bibitem{burrello2021dory}
Alessio Burrello, Angelo Garofalo, Nazareno Bruschi, Giuseppe Tagliavini,
  Davide Rossi, and Francesco Conti.
\newblock Dory: Automatic end-to-end deployment of real-world dnns on low-cost
  iot mcus.
\newblock {\em IEEE Transactions on Computers}, 70(8):1253--1268, 2021.

\bibitem{cai2020tinytl}
Han Cai, Chuang Gan, Ligeng Zhu, and Song Han.
\newblock Tinytl: Reduce activations, not trainable parameters for efficient
  on-device learning.
\newblock {\em arXiv preprint arXiv:2007.11622}, 2020.

\bibitem{cai2019proxylessnas}
Han Cai, Ligeng Zhu, and Song Han.
\newblock {ProxylessNAS: Direct Neural Architecture Search on Target Task and
  Hardware}.
\newblock In {\em ICLR}, 2019.

\bibitem{capotondi2020cmix}
Alessandro Capotondi, Manuele Rusci, Marco Fariselli, and Luca Benini.
\newblock Cmix-nn: Mixed low-precision cnn library for memory-constrained edge
  devices.
\newblock {\em IEEE Transactions on Circuits and Systems II: Express Briefs},
  67(5):871--875, 2020.

\bibitem{chatfield2014return}
Ken Chatfield, Karen Simonyan, Andrea Vedaldi, and Andrew Zisserman.
\newblock Return of the devil in the details: Delving deep into convolutional
  nets.
\newblock In {\em BMVC}, 2014.

\bibitem{chen2015mxnet}
Tianqi Chen, Mu~Li, Yutian Li, Min Lin, Naiyan Wang, Minjie Wang, Tianjun Xiao,
  Bing Xu, Chiyuan Zhang, and Zheng Zhang.
\newblock Mxnet: A flexible and efficient machine learning library for
  heterogeneous distributed systems.
\newblock {\em arXiv preprint arXiv:1512.01274}, 2015.

\bibitem{chen2018tvm}
Tianqi Chen, Thierry Moreau, Ziheng Jiang, Lianmin Zheng, Eddie Yan, Haichen
  Shen, Meghan Cowan, Leyuan Wang, Yuwei Hu, Luis Ceze, et~al.
\newblock $\{$TVM$\}$: An automated end-to-end optimizing compiler for deep
  learning.
\newblock In {\em OSDI}, 2018.

\bibitem{chen2016training}
Tianqi Chen, Bing Xu, Chiyuan Zhang, and Carlos Guestrin.
\newblock Training deep nets with sublinear memory cost.
\newblock {\em arXiv preprint arXiv:1604.06174}, 2016.

\bibitem{choi2018pact}
Jungwook Choi, Zhuo Wang, Swagath Venkataramani, Pierce I-Jen Chuang,
  Vijayalakshmi Srinivasan, and Kailash Gopalakrishnan.
\newblock Pact: Parameterized clipping activation for quantized neural
  networks.
\newblock {\em arXiv preprint arXiv:1805.06085}, 2018.

\bibitem{chowdhery2019visual}
Aakanksha Chowdhery, Pete Warden, Jonathon Shlens, Andrew Howard, and Rocky
  Rhodes.
\newblock Visual wake words dataset.
\newblock {\em arXiv preprint arXiv:1906.05721}, 2019.

\bibitem{cui2018large}
Yin Cui, Yang Song, Chen Sun, Andrew Howard, and Serge Belongie.
\newblock Large scale fine-grained categorization and domain-specific transfer
  learning.
\newblock In {\em CVPR}, 2018.

\bibitem{deng2009imagenet}
Jia Deng, Wei Dong, Richard Socher, Li-Jia Li, Kai Li, and Li~Fei-Fei.
\newblock {ImageNet: A Large-Scale Hierarchical Image Database}.
\newblock In {\em CVPR}, 2009.

\bibitem{donahue2014decaf}
Jeff Donahue, Yangqing Jia, Oriol Vinyals, Judy Hoffman, Ning Zhang, Eric
  Tzeng, and Trevor Darrell.
\newblock Decaf: A deep convolutional activation feature for generic visual
  recognition.
\newblock In {\em ICML}, 2014.

\bibitem{fedorov2019sparse}
Igor Fedorov, Ryan~P Adams, Matthew Mattina, and Paul Whatmough.
\newblock Sparse: Sparse architecture search for cnns on resource-constrained
  microcontrollers.
\newblock In {\em NeurIPS}, 2019.

\bibitem{frankle2020training}
Jonathan Frankle, David~J Schwab, and Ari~S Morcos.
\newblock Training batchnorm and only batchnorm: On the expressive power of
  random features in cnns.
\newblock {\em arXiv preprint arXiv:2003.00152}, 2020.

\bibitem{gan2015devnet}
Chuang Gan, Naiyan Wang, Yi~Yang, Dit-Yan Yeung, and Alex~G Hauptmann.
\newblock Devnet: A deep event network for multimedia event detection and
  evidence recounting.
\newblock In {\em CVPR}, pages 2568--2577, 2015.

\bibitem{goyal2017accurate}
Priya Goyal, Piotr Doll{\'a}r, Ross Girshick, Pieter Noordhuis, Lukasz
  Wesolowski, Aapo Kyrola, Andrew Tulloch, Yangqing Jia, and Kaiming He.
\newblock Accurate, large minibatch sgd: Training imagenet in 1 hour.
\newblock {\em arXiv preprint arXiv:1706.02677}, 2017.

\bibitem{grau2021device}
Marc~Monfort Grau, Roger~Pueyo Centelles, and Felix Freitag.
\newblock On-device training of machine learning models on microcontrollers
  with a look at federated learning.
\newblock In {\em Proceedings of the Conference on Information Technology for
  Social Good}, pages 198--203, 2021.

\bibitem{han2016deep}
Song Han, Huizi Mao, and William~J Dally.
\newblock {Deep Compression: Compressing Deep Neural Networks with Pruning,
  Trained Quantization and Huffman Coding}.
\newblock In {\em ICLR}, 2016.

\bibitem{he2018amc}
Yihui He, Ji~Lin, Zhijian Liu, Hanrui Wang, Li-Jia Li, and Song Han.
\newblock {AMC: AutoML for Model Compression and Acceleration on Mobile
  Devices}.
\newblock In {\em ECCV}, 2018.

\bibitem{he2017channel}
Yihui He, Xiangyu Zhang, and Jian Sun.
\newblock Channel pruning for accelerating very deep neural networks.
\newblock In {\em ICCV}, 2017.

\bibitem{hu2021lora}
Edward Hu, Yelong Shen, Phil Wallis, Zeyuan Allen-Zhu, Yuanzhi Li, Lu~Wang, and
  Weizhu Chen.
\newblock Lora: Low-rank adaptation of large language models, 2021.

\bibitem{ioffe2015batch}
Sergey Ioffe and Christian Szegedy.
\newblock {Batch Normalization: Accelerating Deep Network Training by Reducing
  Internal Covariate Shift}.
\newblock In {\em ICML}, 2015.

\bibitem{jacob2018quantization}
Benoit Jacob, Skirmantas Kligys, Bo~Chen, Menglong Zhu, Matthew Tang, Andrew
  Howard, Hartwig Adam, and Dmitry Kalenichenko.
\newblock Quantization and training of neural networks for efficient
  integer-arithmetic-only inference.
\newblock In {\em Proceedings of the IEEE Conference on Computer Vision and
  Pattern Recognition}, pages 2704--2713, 2018.

\bibitem{jiang2020mnn}
Xiaotang Jiang, Huan Wang, Yiliu Chen, Ziqi Wu, Lichuan Wang, Bin Zou, Yafeng
  Yang, Zongyang Cui, Yu~Cai, Tianhang Yu, et~al.
\newblock Mnn: A universal and efficient inference engine.
\newblock {\em arXiv preprint arXiv:2002.12418}, 2020.

\bibitem{kingma2014adam}
Diederik~P Kingma and Jimmy Ba.
\newblock Adam: A method for stochastic optimization.
\newblock {\em arXiv preprint arXiv:1412.6980}, 2014.

\bibitem{kolesnikov2020big}
Alexander Kolesnikov, Lucas Beyer, Xiaohua Zhai, Joan Puigcerver, Jessica Yung,
  Sylvain Gelly, and Neil Houlsby.
\newblock Big transfer (bit): General visual representation learning.
\newblock In {\em European conference on computer vision}, pages 491--507.
  Springer, 2020.

\bibitem{kornblith2019better}
Simon Kornblith, Jonathon Shlens, and Quoc~V Le.
\newblock Do better imagenet models transfer better?
\newblock In {\em CVPR}, 2019.

\bibitem{krause20133d}
Jonathan Krause, Michael Stark, Jia Deng, and Li~Fei-Fei.
\newblock 3d object representations for fine-grained categorization.
\newblock In {\em Proceedings of the IEEE international conference on computer
  vision workshops}, pages 554--561, 2013.

\bibitem{krizhevsky2009learning}
Alex Krizhevsky, Geoffrey Hinton, et~al.
\newblock Learning multiple layers of features from tiny images.
\newblock 2009.

\bibitem{lai2018cmsis}
Liangzhen Lai, Naveen Suda, and Vikas Chandra.
\newblock Cmsis-nn: Efficient neural network kernels for arm cortex-m cpus.
\newblock {\em arXiv preprint arXiv:1801.06601}, 2018.

\bibitem{langroudi2021tent}
Hamed~F Langroudi, Vedant Karia, Tej Pandit, and Dhireesha Kudithipudi.
\newblock Tent: Efficient quantization of neural networks on the tiny edge with
  tapered fixed point.
\newblock {\em arXiv preprint arXiv:2104.02233}, 2021.

\bibitem{liberis2020mu}
Edgar Liberis, {\L}ukasz Dudziak, and Nicholas~D Lane.
\newblock $\mu$nas: Constrained neural architecture search for
  microcontrollers.
\newblock {\em arXiv preprint arXiv:2010.14246}, 2020.

\bibitem{liberis2019neural}
Edgar Liberis and Nicholas~D Lane.
\newblock Neural networks on microcontrollers: saving memory at inference via
  operator reordering.
\newblock {\em arXiv preprint arXiv:1910.05110}, 2019.

\bibitem{liberis2021differentiable}
Edgar Liberis and Nicholas~D Lane.
\newblock Differentiable network pruning for microcontrollers.
\newblock {\em arXiv preprint arXiv:2110.08350}, 2021.

\bibitem{lin2021mcunetv2}
Ji~Lin, Wei-Ming Chen, Han Cai, Chuang Gan, and Song Han.
\newblock Mcunetv2: Memory-efficient patch-based inference for tiny deep
  learning.
\newblock {\em arXiv preprint arXiv:2110.15352}, 2021.

\bibitem{lin2020mcunet}
Ji~Lin, Wei-Ming Chen, Yujun Lin, John Cohn, Chuang Gan, and Song Han.
\newblock Mcunet: Tiny deep learning on iot devices.
\newblock In {\em NeurIPS}, 2020.

\bibitem{lin2017runtime}
Ji~Lin, Yongming Rao, Jiwen Lu, and Jie Zhou.
\newblock Runtime neural pruning.
\newblock In {\em NeurIPS}, 2017.

\bibitem{liu2019metapruning}
Zechun Liu, Haoyuan Mu, Xiangyu Zhang, Zichao Guo, Xin Yang, Kwang-Ting Cheng,
  and Jian Sun.
\newblock {MetaPruning: Meta Learning for Automatic Neural Network Channel
  Pruning}.
\newblock In {\em ICCV}, 2019.

\bibitem{liu2017learning}
Zhuang Liu, Jianguo Li, Zhiqiang Shen, Gao Huang, Shoumeng Yan, and Changshui
  Zhang.
\newblock Learning efficient convolutional networks through network slimming.
\newblock In {\em ICCV}, 2017.

\bibitem{lyu2021resource}
Bo~Lyu, Hang Yuan, Longfei Lu, and Yunye Zhang.
\newblock Resource-constrained neural architecture search on edge devices.
\newblock {\em IEEE Transactions on Network Science and Engineering}, 2021.

\bibitem{moritz2018ray}
Philipp Moritz, Robert Nishihara, Stephanie Wang, Alexey Tumanov, Richard Liaw,
  Eric Liang, Melih Elibol, Zongheng Yang, William Paul, Michael~I Jordan,
  et~al.
\newblock Ray: A distributed framework for emerging $\{$AI$\}$ applications.
\newblock In {\em 13th USENIX Symposium on Operating Systems Design and
  Implementation (OSDI 18)}, pages 561--577, 2018.

\bibitem{mudrakarta2019k}
Pramod~Kaushik Mudrakarta, Mark Sandler, Andrey Zhmoginov, and Andrew Howard.
\newblock K for the price of 1: Parameter-efficient multi-task and transfer
  learning.
\newblock In {\em ICLR}, 2019.

\bibitem{nilsback2008automated}
Maria-Elena Nilsback and Andrew Zisserman.
\newblock Automated flower classification over a large number of classes.
\newblock In {\em 2008 Sixth Indian Conference on Computer Vision, Graphics \&
  Image Processing}, pages 722--729. IEEE, 2008.

\bibitem{parkhi2012cats}
Omkar~M Parkhi, Andrea Vedaldi, Andrew Zisserman, and CV~Jawahar.
\newblock Cats and dogs.
\newblock In {\em 2012 IEEE conference on computer vision and pattern
  recognition}, pages 3498--3505. IEEE, 2012.

\bibitem{pytorch2019}
Adam Paszke, Sam Gross, Francisco Massa, Adam Lerer, James Bradbury, Gregory
  Chanan, Trevor Killeen, Zeming Lin, Natalia Gimelshein, Luca Antiga, et~al.
\newblock Pytorch: An imperative style, high-performance deep learning library.
\newblock {\em Advances in neural information processing systems}, 32, 2019.

\bibitem{patil2022poet}
Shishir~G Patil, Paras Jain, Prabal Dutta, Ion Stoica, and Joseph Gonzalez.
\newblock Poet: Training neural networks on tiny devices with integrated
  rematerialization and paging.
\newblock In {\em International Conference on Machine Learning}, pages
  17573--17583. PMLR, 2022.

\bibitem{rastegari2016xnor}
Mohammad Rastegari, Vicente Ordonez, Joseph Redmon, and Ali Farhadi.
\newblock Xnor-net: Imagenet classification using binary convolutional neural
  networks.
\newblock In {\em ECCV}, 2016.

\bibitem{ren2021tinyol}
Haoyu Ren, Darko Anicic, and Thomas~A Runkler.
\newblock Tinyol: Tinyml with online-learning on microcontrollers.
\newblock In {\em 2021 International Joint Conference on Neural Networks
  (IJCNN)}, pages 1--8. IEEE, 2021.

\bibitem{rusci2019memory}
Manuele Rusci, Alessandro Capotondi, and Luca Benini.
\newblock Memory-driven mixed low precision quantization for enabling deep
  network inference on microcontrollers.
\newblock In {\em MLSys}, 2020.

\bibitem{sandler2018mobilenetv2}
Mark Sandler, Andrew Howard, Menglong Zhu, Andrey Zhmoginov, and Liang-Chieh
  Chen.
\newblock {MobileNetV2: Inverted Residuals and Linear Bottlenecks}.
\newblock In {\em CVPR}, 2018.

\bibitem{sharif2014cnn}
Ali Sharif~Razavian, Hossein Azizpour, Josephine Sullivan, and Stefan Carlsson.
\newblock Cnn features off-the-shelf: an astounding baseline for recognition.
\newblock In {\em CVPR Workshops}, 2014.

\bibitem{sudharsan2021globe2train}
Bharath Sudharsan, John~G Breslin, and Muhammad~Intizar Ali.
\newblock Globe2train: A framework for distributed ml model training using iot
  devices across the globe.
\newblock In {\em 2021 IEEE SmartWorld, Ubiquitous Intelligence \& Computing,
  Advanced \& Trusted Computing, Scalable Computing \& Communications, Internet
  of People and Smart City Innovation (SmartWorld/SCALCOM/UIC/ATC/IOP/SCI)},
  pages 107--114. IEEE, 2021.

\bibitem{sudharsan2021train}
Bharath Sudharsan, Piyush Yadav, John~G Breslin, and Muhammad~Intizar Ali.
\newblock Train++: An incremental ml model training algorithm to create
  self-learning iot devices.
\newblock In {\em 2021 IEEE SmartWorld, Ubiquitous Intelligence \& Computing,
  Advanced \& Trusted Computing, Scalable Computing \& Communications, Internet
  of People and Smart City Innovation (SmartWorld/SCALCOM/UIC/ATC/IOP/SCI)},
  pages 97--106. IEEE, 2021.

\bibitem{tan2019mnasnet}
Mingxing Tan, Bo~Chen, Ruoming Pang, Vijay Vasudevan, Mark Sandler, Andrew
  Howard, and Quoc~V Le.
\newblock {MnasNet: Platform-Aware Neural Architecture Search for Mobile}.
\newblock In {\em CVPR}, 2019.

\bibitem{vaswani2017attention}
Ashish Vaswani, Noam Shazeer, Niki Parmar, Jakob Uszkoreit, Llion Jones,
  Aidan~N Gomez, Lukasz Kaiser, and Illia Polosukhin.
\newblock Attention is all you need.
\newblock {\em arXiv preprint arXiv:1706.03762}, 2017.

\bibitem{wang2019haq}
Kuan Wang, Zhijian Liu, Yujun Lin, Ji~Lin, and Song Han.
\newblock {HAQ: Hardware-Aware Automated Quantization with Mixed Precision}.
\newblock In {\em CVPR}, 2019.

\bibitem{cub}
Peter Welinder, Steve Branson, Takeshi Mita, Catherine Wah, Florian Schroff,
  Serge Belongie, and Pietro Perona.
\newblock Caltech-ucsd birds 200.
\newblock Technical Report CNS-TR-201, Caltech, 2010.

\bibitem{you2017large}
Yang You, Igor Gitman, and Boris Ginsburg.
\newblock Large batch training of convolutional networks.
\newblock {\em arXiv preprint arXiv:1708.03888}, 2017.

\bibitem{yu2017scalpel}
Jiecao Yu, Andrew Lukefahr, David Palframan, Ganesh Dasika, Reetuparna Das, and
  Scott Mahlke.
\newblock Scalpel: Customizing dnn pruning to the underlying hardware
  parallelism.
\newblock {\em ACM SIGARCH Computer Architecture News}, 45(2):548--560, 2017.

\bibitem{zaken21bitfit}
Elad~Ben Zaken, Shauli Ravfogel, and Yoav Goldberg.
\newblock Bitfit: Simple parameter-efficient fine-tuning for transformer-based
  masked language-models.
\newblock {\em CoRR}, abs/2106.10199, 2021.

\bibitem{zoph2017neural}
Barret Zoph and Quoc~V Le.
\newblock {Neural Architecture Search with Reinforcement Learning}.
\newblock In {\em ICLR}, 2017.

\bibitem{zoph2018learning}
Barret Zoph, Vijay Vasudevan, Jonathon Shlens, and Quoc~V. Le.
\newblock {Learning Transferable Architectures for Scalable Image Recognition}.
\newblock In {\em CVPR}, 2018.

\end{thebibliography}

\newpage
\appendix
\section{Video Demo}
\label{sec:video_demo}
We prepared a video demo showing that we can deploy our framework to a microcontroller (STM32F746, 320KB SRAM, 1MB Flash) to enable on-device learning. We adapt the MCUNet model (pre-trained on ImageNet) to classify whether there is a person in front of the camera or not. The training leads to decent accuracy within the tight memory budget. Please find the demo here: \href{https://youtu.be/XaDCO8YtmBw}{https://youtu.be/XaDCO8YtmBw}.

\begin{figure*}[h]
    \centering
    \includegraphics[width=0.4\textwidth]{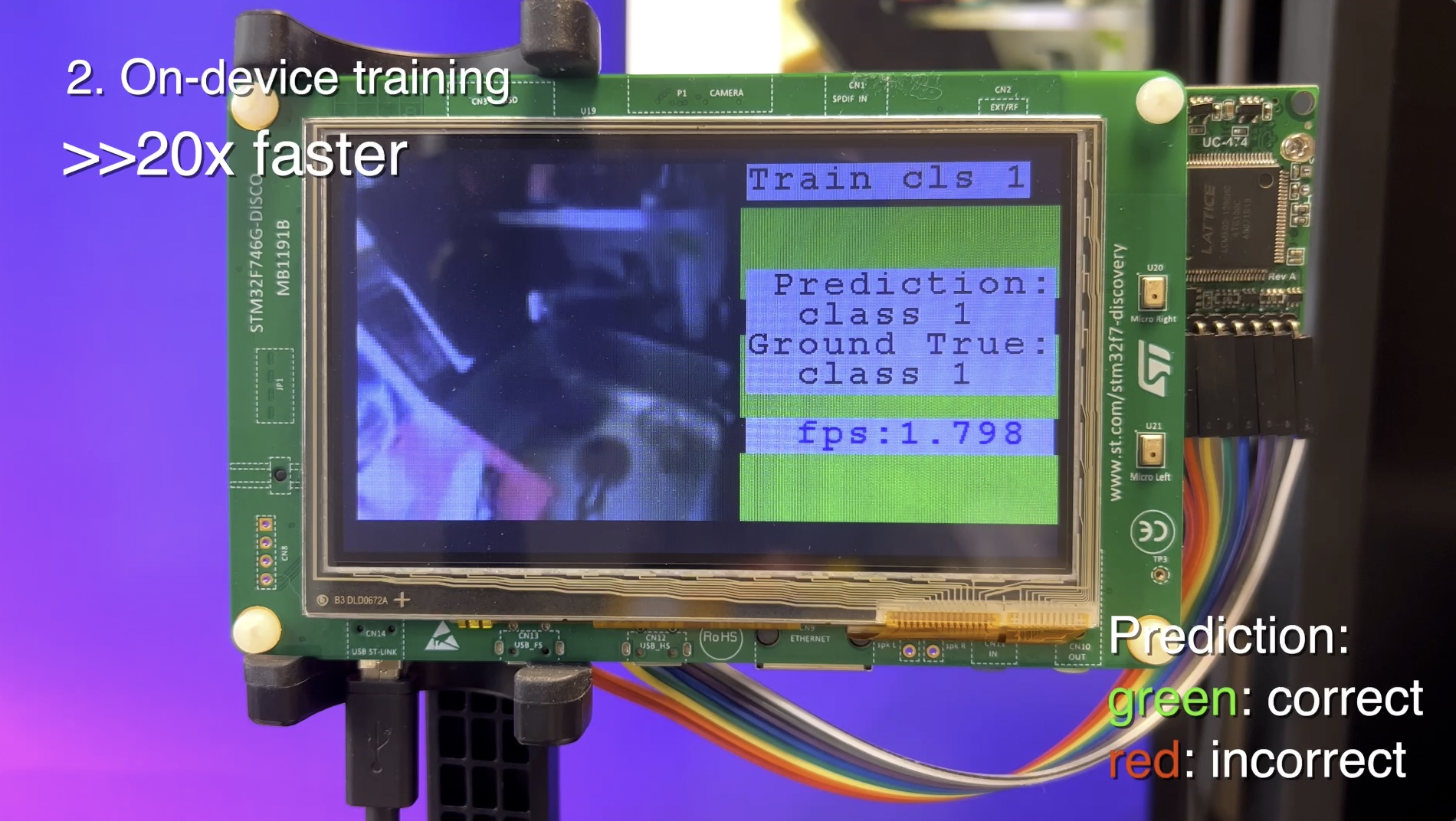}
    \includegraphics[width=0.4\textwidth]{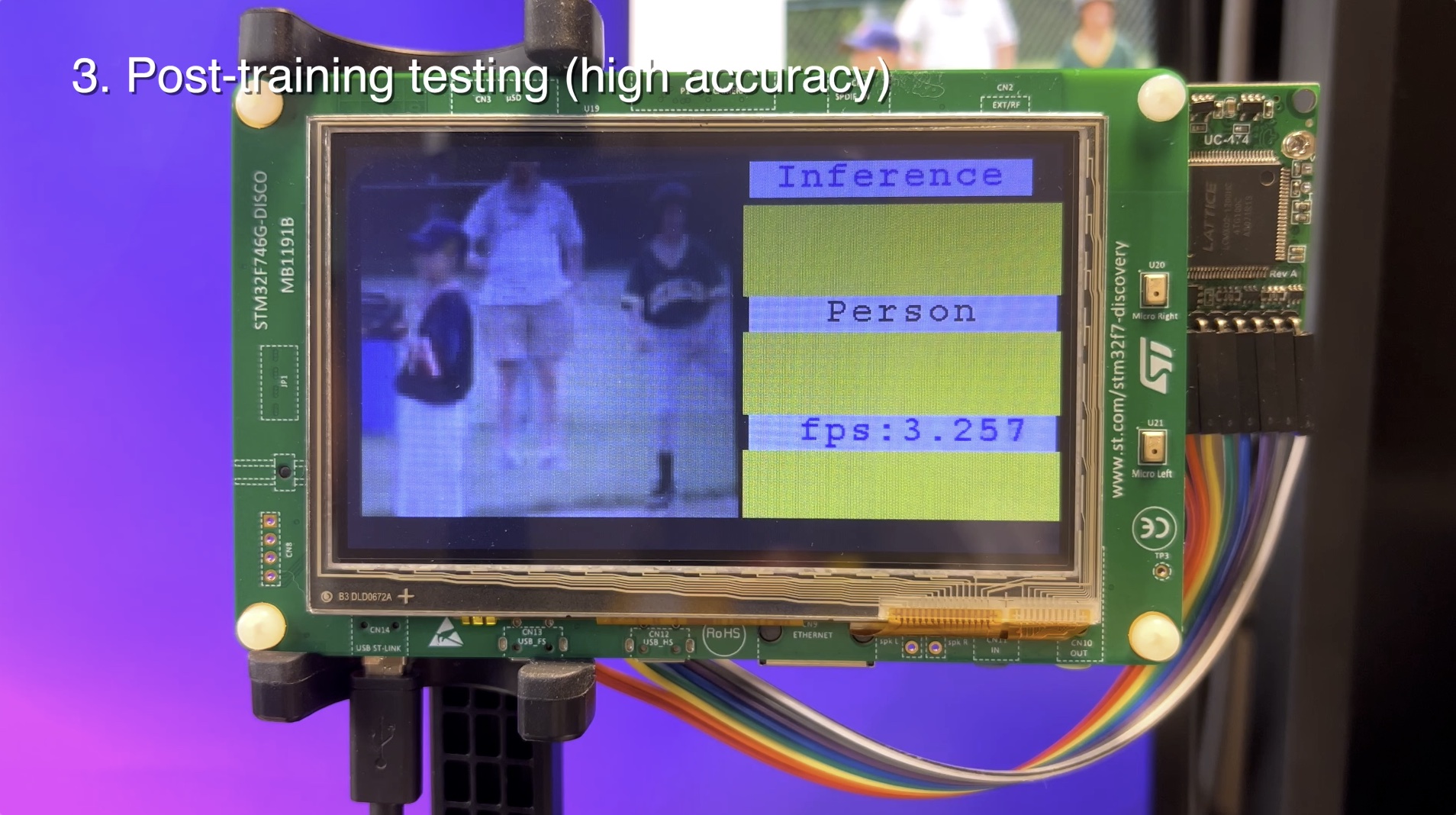}
    \caption{A screenshot of our video demo.}
    \label{fig:demo_screenshot}
\end{figure*}

The training is performed with 100 sample images from the VWW dataset~\cite{chowdhery2019visual} fed through the camera (50 positive and 50 negative). The total (pure) training throughput for the pipeline (including overheads like camera IO) is shown in the Figure.~\ref{fig:demo_screenshot}. The total training time would be around minutes. This is quite affordable for tiny on-device learning applications.

\section{Variance of Different Runs}

We notice that the variance of different runs is quite small in our experiments. %
Here we provide detailed information about the variance. 

Firstly, if we use the same random seed for the data loader, we will get \emph{exactly the same} results for multiple runs. The weight quantization process after each iteration (almost) eliminates the non-determinism from GPU training\footnote{\url{https://developer.download.nvidia.com/video/gputechconf/gtc/2019/presentation/s9911-determinism-in-deep-learning.pdf}}. 
Therefore, we study the randomness from different random seeds in \emph{data shuffling}.
Here we provide the results of 3 runs in Table~\ref{tab:std_study} to show the variance. We train the MobileNetV2-w0.35 model with the sparse update scheme (searched under 100KB analytic memory usage) 3 times independently. We find the variance is very small, especially when we report the average accuracy (for most of our results): the standard derivation is only $\pm0.07\%$.

\begin{table}[H]
    \setlength{\tabcolsep}{6pt}
    \caption{The variance between different runs is small, especially when we report the average accuracy (only $\pm0.07\%$). Results obtained by training MobileNetV2-w0.35 for three times using the sparse update scheme searched under 100KB analytic memory constraint. 
    }
    \label{tab:std_study}
    \centering
    \small{
     \begin{tabular}{lccccccccc}
    \toprule
  \multirow{2}{*}{Runs} &  \multicolumn{8}{c}{Accuracy (\%)} & \multirow{2}{*}{\shortstack{Avg\\Acc.}} \\ \cmidrule(lr){2-9}
 & Cars & CF10 & CF100 & CUB & Flowers & Food & Pets & VWW \\
    \midrule
 run1 & 51.59 & 87.03 & 63.89 & 54.14 & 85.95 & 62.28 & 77.84 & 88.34 & 71.38 \\ 
 run2 & 52.87 & 86.8 & 63.81 & 54.87 & 85.30 & 62.45 & 77.30 & 88.65 & 71.50 \\
 run3 & 52.49 & 87.13 & 63.80 & 55.16 & 85.35 & 61.99 & 77.08 & 88.21 & 71.40 \\ \midrule
 mean & 52.32 & 86.99 & 63.83 & 54.72 & 85.53 & 62.24 & 77.41 & 88.40 & \textbf{71.43} \\
 $\pm$std & $\pm$0.66 & $\pm$0.17 & $\pm$0.05 & $\pm$0.52 & $\pm$0.36 & $\pm$0.23 & $\pm$0.39 & $\pm$0.22 & $\mathbf{\pm}$\textbf{0.07} \\
    \bottomrule
     \end{tabular}
     }
\end{table}

\section{Training Setups \& Discussions}
\label{sec:training_details}

In this section, we introduce detailed training setups and discuss the reasons that lead to several design choices.

We used SGD optimizer+QAS for training. We set weight decay as 0 since we observed no over-fitting during experiments. This is also a common choice in transfer learning~\cite{kolesnikov2020big}.
We find the initial learning rate significantly affects the accuracy, so we extensively tuned the learning rate for each run to report the best accuracy. 
We used cosine learning rate decay and performed warm-up~\cite{goyal2017accurate} for 1 epoch on VWW and 5 epochs on other datasets. 
We used Ray~\cite{moritz2018ray} for experiment launching and hyper-parameter tuning. %

\paragraph{Data type of the classifier. } 
During transfer learning, we usually need to randomly initialize the classifiers (or add some classes) for novel categories. 
Although the backbone is fully quantized for efficiency, we find that using a floating-point classifier is essential for transfer learning performance. Using a floating-point classifier is also cost-economical since the classifier consists of a very small part of the model size (0.3\% for 10 classes).

We compare the results of the quantized classifier and floating-point classifier in Table~\ref{tab:fc_study}. We update the last two blocks of the MCUNet model with SGD-M optimizer and QAS to measure the downstream accuracy. We find that keeping the classifier as floating-point significantly improves the downstream accuracy by 2.3\% (on average) at a marginal overhead. \emph{Therefore, we use floating-point for the classifier by default.}

\begin{table}[H]
    \caption{Keeping the classifier as floating-point greatly improves the downstream accuracy. 
    }
    \label{tab:fc_study}
    \centering
    \small{
     \begin{tabular}{cccccccccc}
    \toprule
  \multirow{2}{*}{\shortstack{\texttt{fp32}\\classifier}}&  \multicolumn{8}{c}{Accuracy (\%)} & \multirow{2}{*}{\shortstack{Avg\\Acc.}} \\ \cmidrule(lr){2-9}
 & Cars & CF10 & CF100 & CUB & Flowers & Food & Pets & VWW \\ \midrule
 \ding{55}  & 50.8 & 86.1 & 62.7 & 56.8 & 82.5 & 61.7 & 80.8 & 87.8 & 71.2  \\
 \ding{51} & 55.2 & 86.9 & 64.6 & 57.8 & 89.1 & 64.4 & 80.9 & 89.3 & \textbf{73.5} \\
    \bottomrule
     \end{tabular}
     }
\end{table}

\myparagraph{Single-batch training \& momentum.}
For on-device training on microcontrollers, we can only fit batch size 1 due to the tight memory constraint. However, single-batch training has very low efficiency when simulated on GPUs since it cannot leverage the hardware parallelism, making experiments slow. We study the performance gap between single-batch training and normal-batch training (batch size 128) to see if we can use the latter as an approximation.

We compare the results of different batch sizes in Table~\ref{tab:single_batch}, with and without momentum. Due to the extremely low efficiency of single-batch training, we only report results on datasets of a smaller size. 
We used SGD+QAS as the optimizer and updated the last two blocks of the MCUNet~\cite{lin2020mcunet} model. We extensively tuned the initial learning rate to report the best results.

\definecolor{this_gray}{rgb}{0.8,0.8,0.8}
\begin{table}[H]
    \caption{Momentum helps transfer learning with batch size 128, but not with batch size 1; without momentum, we can use the normal-batch training results as an approximation for single-batch training.
    Results obtained by updating the last two blocks of MCUNet~\cite{lin2020mcunet} with SGD+QARS.
    }
    \label{tab:single_batch}
    \centering
    \small{
     \begin{tabular}{llcccccccc}
    \toprule
 \multirow{2}{*}{Batch size} & \multirow{2}{*}{Momentum} & \multirow{2}{*}{Mem Cost} & \multicolumn{5}{c}{Accuracy (\%)} & \multirow{2}{*}{\shortstack{Avg\\Acc.}} \\ \cmidrule(lr){4-8}
 & & & Cars &  CUB & Flowers & Pets & VWW \\
\midrule
 \multirow{2}{*}{128 \textcolor{this_gray}{(GPU simulate)}} & 0.9 & 2$\times$ & 55.2 & 57.8 & 89.1 & 80.9  & 89.3 & 74.4 \\ 
 & 0 &  1$\times$ & 47.8 & 57.2 & 87.3 & 80.8 & 88.8&  72.4  \\ \midrule 
\multirow{2}{*}{1 \textcolor{this_gray}{(tinyML)} } & 0.9 &  2$\times$& 47.8 &  54.8 & 88.5 & 80.5 & 86.2  & 71.5 \\ 
& 0 &   1$\times$ &51.1 &  56.2 & 88.7 & 79.3 & 86.0 &  72.3 \\ 
    \bottomrule
     \end{tabular}
     }
\end{table}

We can make two observations:
\begin{enumerate}
    \item Firstly, momentum helps optimization for normal-batch training as expected (average accuracy 74.4\% \vs 72.4\%). However, it actually makes the accuracy slightly worse for the single-batch setting (71.5\% \vs 72.3\%). Since using momentum will double the memory requirement for updating parameters (assume we can safely quantize momentum buffer; otherwise the memory usage will be 5$\times$ larger), we will not use momentum for tinyML on-device learning.
    \item Without momentum, normal-batch training, and single-batch training lead to a similar average accuracy (72.4\% \vs 72.3\%), allowing us to use batched training results for evaluation.
\end{enumerate}

Given the above observation, \emph{we report the results of batched training without momentum by default}, unless otherwise stated.

\myparagraph{Gradient accumulation.}
With the above training setting, we can get a similar average accuracy compared to actual on-device training on microcontrollers. The reported accuracy on each dataset is quite close to the real on-device accuracy, with \emph{only one exception}: the VWW dataset, where the accuracy is 2.5\% lower. This is because VWW only has two categories (binary classification), so the information from each label is small, leading to unstable gradients. For the cases where the number of categories is small, we can add gradient accumulation to make the update more stable. We show the comparison of adapting the pre-trained MCUNet model in  Table~\ref{tab:grad_accumulate}. The practice closes the accuracy gap at a small extra memory cost (11\%), allowing us to get 89.1\% top-1 accuracy within 256KB memory usage.

To provide a clear comparison, we \emph{do not} apply gradient accumulation in our experiments except for this comparison. 

\begin{table}[H]
    \caption{Gradient accumulation helps the optimization on datasets with a small category number. Numbers obtained by training with batch size 1, the same setting as on microcontrollers. 
    }
    \label{tab:grad_accumulate}
    \centering
    \small{
     \begin{tabular}{lccc}
    \toprule
 model & accumulate grad & SRAM & VWW accuracy \\ \midrule
 \multirow{2}{*}{MCUNet-5FPS} &  \ding{55}  & 160KB & 86.6\% \\
&  \ding{51} & 188KB & 89.1\% \\
    \bottomrule
     \end{tabular}
     }
\end{table}

\section{Evolutionary Search \vs Random Search}

\label{sec:evo_vs_random}

\begin{figure*}[h]
    \centering
     \includegraphics[width=0.5\textwidth]{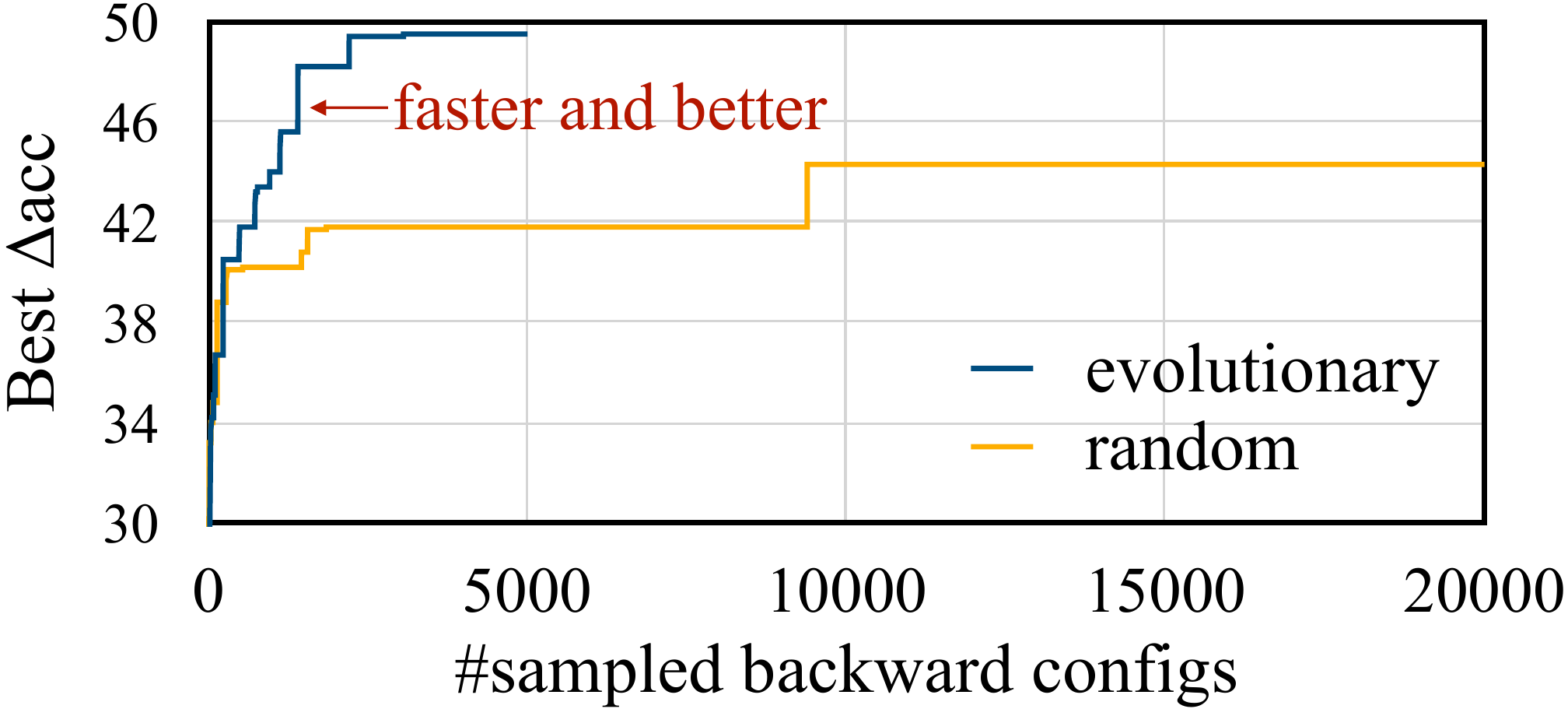}
    \caption{Evolutionary search has a better sample efficiency and leads to a better final result compared with random search when optimizing sparse update schemes. 
    }
    \label{fig:evolutionary_vs_random}
\end{figure*}

We find that evolutionary search can efficiently explore the search space to find a good sparse update scheme given a memory constraint. Here we provide the comparison between evolutionary search and random search in Figure~\ref{fig:evolutionary_vs_random}. We collect the curves when searching for an update scheme of the MCUNet-5FPS~\cite{lin2020mcunet} model under 100KB memory constraint (analytic). We find that evolutionary search has a much better sample efficiency and can find a better final solution (higher sum of $\Delta\text{acc}$) compared to random search. The search process is quite efficient: we can search for a sparse update scheme within 10 minutes based on the contribution information. Note that we use the \emph{same} update scheme for all downstream datasets. 

\section{Amount of Compute}

To evaluate the performance of different training schemes, we simulate the training on GPUs to measure the average accuracy on 8 downstream datasets. 
Thanks to the small model size (for the tinyML setting) and the small dataset size, the training cost for each scheme is quite modest: it only takes \textbf{3.2 GPU hours} for training on all 8 downstream datasets (cost for one run; do not consider hyper-parameter tuning). 

For the pre-training on ImageNet~\cite{deng2009imagenet}, it takes about \textbf{31.5 GPU hours} (300 epochs). Note that we only need to pre-train each model \emph{once}.

We performed training with NVIDIA GeForce RTX 3090 GPUs.

\section{More Contribution Analysis Results}
\label{sec:more_contribution}
\begin{figure*}[h]
    \centering
     \includegraphics[width=\textwidth]{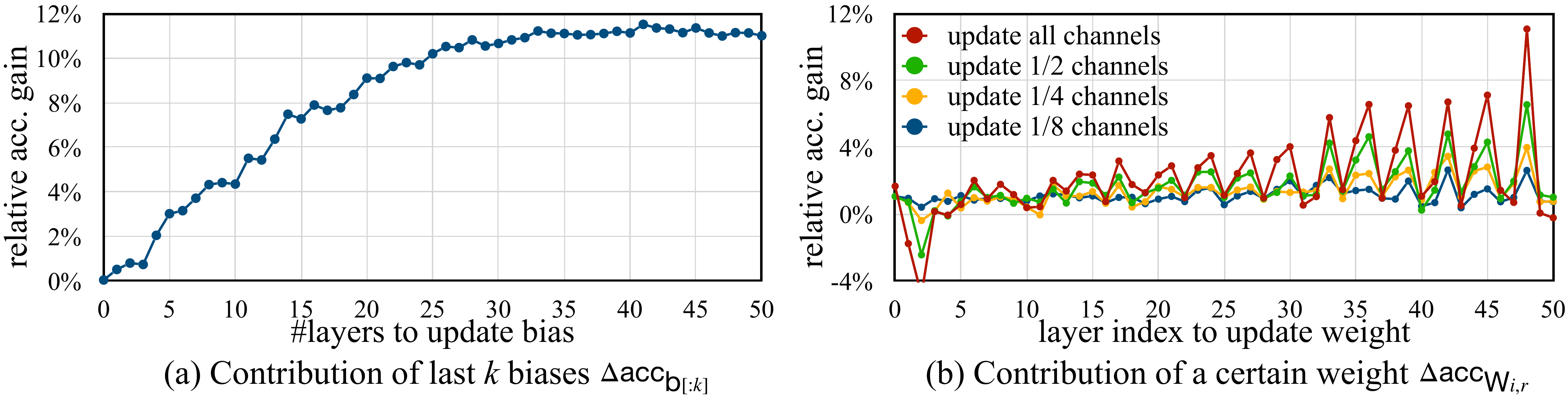}
    \caption{Contribution analysis of updating biases and weights for MobileNetV2-w0.35~\cite{sandler2018mobilenetv2}.
    }
    \label{fig:sensitivity_curve_mbv2}
\end{figure*}

\begin{figure*}[h]
    \centering
     \includegraphics[width=\textwidth]{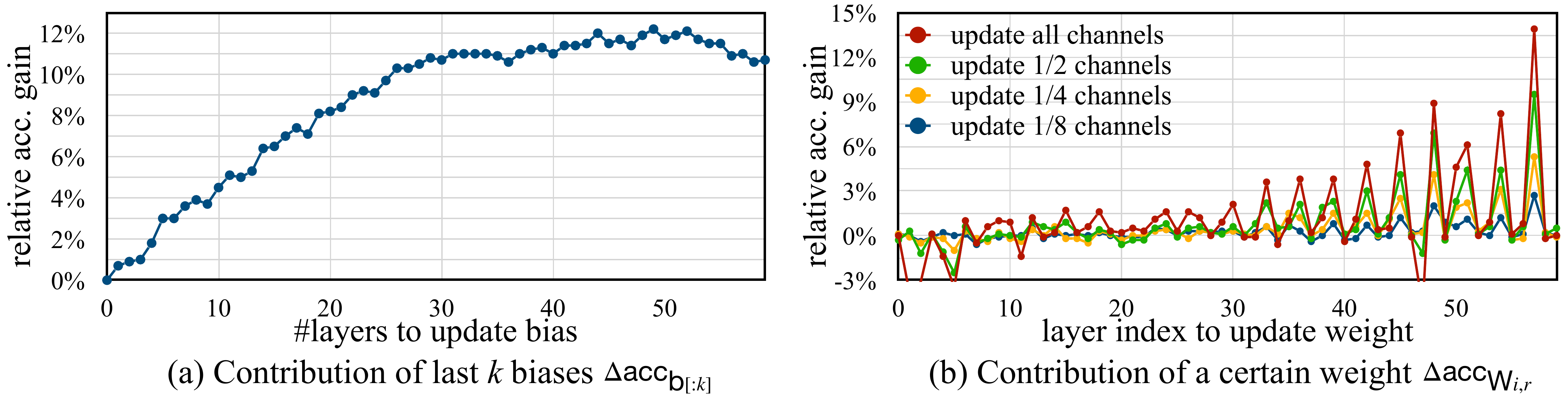}
    \caption{Contribution analysis of updating biases and weights for ProxylessNAS-w0.3~\cite{cai2019proxylessnas}.
    }
    \label{fig:sensitivity_curve_proxyless}
\end{figure*}

Here we provide the contribution analysis results of the MobileNetV2-w0.35~\cite{sandler2018mobilenetv2} and ProxylessNAS-w0.3~\cite{cai2019proxylessnas} on the Cars dataset~\cite{krause20133d} (Figure~\ref{fig:sensitivity_curve_mbv2} and \ref{fig:sensitivity_curve_proxyless}). The pattern is similar to the one from the MCUNet model: the later layers contribute to the accuracy improvement more; within each block, the first point-wise convolutional layer contributes to the accuracy improvement the most.

\newpage
\section{Other Partial Update Methods That Did Not Work}

During our experiments, we also considered other efficient partial update methods (apart from sparse layer/tensor update) but they did not work well. Here are a few methods we tried but failed:

\paragraph{1. Low-rank update.} LoRA~\cite{hu2021lora} aims to adapt a model by adding a low-rank decomposed weight to each of the original weight matrix. It is designed for adapting large language models, but could potentially be applied here. Specifically, LoRA freezes the original weight $\mathbf{W}\in\mathds{R}^{c\times c}$ but trains a small $\Delta\mathbf{W} = \mathbf{MN}$, where $\mathbf{M}\in\mathds{R}^{c\times c'}, \mathbf{N}\in\mathds{R}^{c'\times c}, c'<<c$. The low-rank decomposed $\Delta\mathbf{W}$ has much fewer parameters compared to $\mathbf{W}$. After training, we can merge $\Delta\mathbf{W}$ so that no extra computation is incurred: $\mathbf{y} = (\mathbf{W} + \Delta\mathbf{W})\mathbf{x}$. However, such method does not work in our case:
\begin{enumerate}
    \item  The weights are quantized in our models. If we merge $\Delta\mathbf{W}$ and $\mathbf{W}$, we will produce a new weight $\mathbf{W}' = \Delta\mathbf{W} + \mathbf{W}$ that has the same size as $\mathbf{W}$, taking up a large space on the SRAM (that is why we need the sparse tensor update). 
    \item Even if we can tolerate the extra memory overhead by running $\mathbf{y} = \mathbf{W}\mathbf{x} + \Delta\mathbf{W}\mathbf{x}$, the $\Delta\mathbf{W}$ is randomly initialized and we empirically find that it is difficult to update a quantized weight from scratch, leading to worse performance.
\end{enumerate}

\paragraph{2. Replacing convolutions with lighter alternatives.}
As shown in the contribution curves (Figure 4 in the main paper, Figure~\ref{fig:sensitivity_curve_mbv2}, and Figure~\ref{fig:sensitivity_curve_proxyless}), the first point-wise convolutional layer in each block has the highest contribution to accuracy. We tried replacing the first point-wise convolutional layer with a lighter alternative, like grouped convolutions. However, although such replacement greatly reduces the cost to update the layers, it also hinders transfer learning accuracy significantly. Therefore, we did not choose to use such modification. It also involves extra complexity by changing model architectures, which is not desired.

\section{Changelog}
\paragraph{v1} Initial preprint release.
\paragraph{v2} Fix a typo in Equation~\ref{eq:qas}.  
\paragraph{v3} Camera-ready version.
\paragraph{v4} Update project and demo links.

\end{document}